\pdfoutput=1

\documentclass[11pt]{article}

\usepackage[final]{acl}

\usepackage{times}
\usepackage{lscape}
\usepackage{svg}
\usepackage{adjustbox}
\usepackage{tabularx} 
\usepackage{makecell} 
\usepackage{latexsym}
\usepackage{booktabs} 

\usepackage{amsmath,amssymb,amsfonts}
\usepackage{hyperref}
\usepackage{subcaption}
\usepackage{enumitem}       
\usepackage{array}          
\usepackage{geometry}       
\usepackage{xspace}
\usepackage{multicol}
\usepackage{multirow}
\usepackage{subcaption}
\usepackage{tcolorbox}
\usepackage{lscape}
\usepackage{algorithm}
\usepackage{algpseudocode}
\usepackage[T1]{fontenc}

\usepackage[utf8]{inputenc}

\usepackage{microtype}
\usepackage{inconsolata}

\usepackage{graphicx}
\usepackage{comment}

\newcommand{\nummodels}{14}
\newcommand{\numinsights}{three}

\title{Lost in Transcription, Found in Distribution Shift: Demystifying Hallucination in Speech Foundation Models}

\author{%
  Hanin Atwany$^{2}$\thanks{Equal contributing first authors. Correspondence: \texttt{hanin.atwany@mbzuai.ac.ae, abdulw@cs.cmu.edu}} \,
  Abdul Waheed$^{1}$\footnotemark[1] \,
  Rita Singh$^{1}$ \,
  Monojit Choudhury$^{2}$\,
  Bhiksha Raj$^{1}$ \\[1ex]
  $^{1}$Carnegie Mellon University \quad 
  $^{2}$MBZUAI
}

\begin{document}
\maketitle
\begin{abstract}

Speech foundation models trained at a massive scale, both in terms of model and data size, result in robust systems capable of performing multiple speech tasks, including automatic speech recognition (ASR). These models transcend language and domain barriers, yet effectively measuring their performance remains a challenge. Traditional metrics like word error rate (WER) and character error rate (CER) are commonly used to evaluate ASR performance but often fail to reflect transcription quality in critical contexts, particularly when detecting fabricated outputs. This phenomenon, known as hallucination, is especially concerning in high-stakes domains such as healthcare, legal, and aviation, where errors can have severe consequences. In our work, we address this gap by investigating hallucination in ASR models. We examine how factors such as distribution shifts, model size, and model architecture influence the hallucination error rate (HER), a metric we introduce to quantify hallucinations. Our analysis of over 20 ASR models reveals \numinsights~key insights: (1) High WERs can mask low hallucination rates, while low WERs may conceal dangerous hallucinations. (2) Synthetic noise, both adversarial and common perturbations like white noise, pitch shift, and time stretching, increase HER. (3) Distribution shift correlates strongly with HER ($\alpha = 0.91$). Our findings highlight the importance of incorporating HER alongside traditional metrics like WER to better assess ASR model performance, particularly in high-stakes domains.

%
\end{abstract}

\section{Introduction}\label{sec:introduction}

Automatic Speech Recognition (ASR) systems have become fundamental to various applications, including personal assistants, automated customer service, and transcription tools used in fields such as education, healthcare, and law~\cite{zhang2023intelligent, adedeji2024sound}. These systems have seen remarkable improvements in recent years~\cite{arriaga2024evaluation,radford2022robustspeechrecognitionlargescale, communication2023seamlessm4tmassivelymultilingual}, with state-of-the-art models demonstrating their capabilities across diverse datasets and languages~\cite{shakhadri2025samba}. However, the evaluation of ASR performance remains largely dependent on word and character error rate (WER/CER). The primary limitation of WER and CER is their dependence on token-level overlapping, which focuses on matching individual words or characters without considering the overall semantic aspect of the transcription. This could result in misleading evaluations, as a high WER/CER does not necessarily indicate poor outputs in all cases.
In addition, these metrics fall short in capturing more subtle semantic failures which are not typically caught without human verification, such as hallucinations.

Hallucinations in ASR systems mirror perceptual experiences in neuroscience—plausible outputs generated without grounding in input stimuli~\cite{american2013diagnostic,zmigrod2016neural}, deviating \textit{phonetically} or \textit{semantically} from source speech~\cite{ji2023survey}. Like natural neural perceptions, ASR hallucinations arise when models prioritize distributional patterns over fidelity to audio input, fabricating text unlinked to reference content~\cite{hare2021hallucinations}. These errors are uniquely hazardous in high-stakes domains~\cite{williamson2024era}, as they evade WER/CER detection while distorting meaning, similar to how clinical hallucinations disconnect from reality.

Hallucination in domains such as medical and legal can have serious consequences, including life-threatening outcomes and distorted testimonies or contracts, and may disproportionately affect marginalized groups~\cite{faithful_ai_in_medicine, vishwanath2024faithfulness, mujtaba2024lost, sperber2020consistent, Koenecke_2024}. Hence, detecting and mitigating hallucination is crucial for ensuring the reliability of ASR systems in sensitive environments.

The existing literature on hallucination detection in ASR models is confined to a single model~\cite{frieske2024hallucinations, serai2021hallucinationspeechrecognitionerrors, Koenecke_2024, barański2025investigationwhisperasrhallucinations, ji2023survey} or test setting~\cite{kim2024automatic}, highlighting a significant research gap for a systematic investigation across different supervision paradigms, test domains, and conditions.

In this work, we address this critical research gap and make the following key contributions:

\begin{itemize}
    \item We conduct a thorough evaluation of ASR models across various setups, consisting of both synthetic and natural shifts from training to test distributions.
    \item We introduce an LLM-based error detection framework that classifies ASR outputs into different types of errors including hallucination errors through context-aware assessments.
    \item We provide an in-depth analysis of hallucination phenomena in ASR models, exploring the impact of domain-specific data, model architectures, and training paradigms, and offer valuable insights into the relationship between model size, type, and hallucination frequency.
    \item To validate our hallucination detection method, we compare our LLM-based hallucination detection pipeline with human evaluations and heuristic approaches, demonstrating that the LLM evaluation closely aligns with human judgments and other LLM-based assessments, unlike the heuristic-based approach.
\end{itemize}

The significance of this work lies in how it redefines the evaluation and improvement of ASR systems. By emphasizing hallucination detection, we aim to enhance the reliability of ASR models, particularly in domains where accuracy and precision are non-negotiable.

\noindent\textbf{Outline.} In Section~\ref{sec:related_work}, we review prior work related to ASR evaluation and hallucination detection. Section~\ref{sec:methodology} outlines our proposed methodology. Section~\ref{sec:experiments} presents our experimental setup. Finally, Section~\ref{sec:results} discusses the results and implications of our findings, outlining directions for future research. We conclude our work in Section~\ref{sec:conclusion} and provide limitations in Section~\ref{sec:limitations}.

\begin{table*}[ht]
\resizebox{\textwidth}{!}{%
\begin{tabular}{p{4cm}p{5cm}p{5cm}p{1cm}p{1cm}p{1cm}p{1cm}p{3cm}}
\toprule
\multirow{2}{*}{Model} &
\multirow{2}{*}{Reference} &
\multirow{2}{*}{Hypothesis} &
 WER & \multicolumn{3}{c}{Hallucination} & \multirow{2}{*}{Setting} \\
& & &  & Heuristic & Human & LLM & \\ \midrule
whisper-small & hungry action hippos fruit & Humm reaction in hippos fruit? & 75 & F & F & F & Home Environment \\ 
hf-seamless-m4t-large & probably i i had asthma & The song is Erasmus. & 100 & F & T & T & Primock57 (medical) \\
hf-seamless-m4t-large & it can't be done & I'm going to start with the first one. & 200 & F & T & T & Superemecourt (Legal) \\
whisper-medium & patel para thirty eight page three hundred and fifty five & How much is the tail? & 100 & F & T & T & Legal \\
hubert & god i'm forty five & god he whisper i'm forty five & 40 & F & T & T & Primock57 (medical)\\
hubert & yeah & in a handsome coat & 400 & F & T & T & AMI (meetings) \\
wav2vec & today yeah & she did yes & 150 & F & T & T & Primock57 (medical) \\
wav2vec2 & another dog secretary show & i love her dog's secretary show & 100 & F & T & T & BERSt (Noisy) \\ 
\bottomrule
\end{tabular}
}
\caption{Examples showing hallucination detection by different methods across domains and models.}
\label{tab:examples}
\end{table*}

\section{Related Work}\label{sec:related_work}
The use of ASR systems in high-stakes domains, including healthcare~\cite{afonja2024performantasrmodelsmedical, huh2023improvingmedicalspeechtotextaccuracy, adedeji2024sound, sunder2022buildingasrerrorrobust}, legal proceedings~\cite{saadany2022bettertranscriptionuksupreme, garneau2024statecommercialautomaticfrench}, and finance~\cite{Del_Rio_2021, 10389617}, has heightened the necessity for ensuring their robustness. Conventionally, the performance of these models is assessed using metrics such as WER and CER~\cite{serai2022hallucination}.~\citet{szymanski-etal-2023-arent, sasindran2024semascore} They show that when these metrics are used in isolation, they exhibit notable limitations.

Recent studies have extensively investigated hallucination in text generated by large language models (LLMs), identifying it as a prevalent phenomenon~\cite{Huang_2025, bai2024hallucinationmultimodallargelanguage, yao2023llm, jiang2024surveylargelanguagemodel,maynez2020faithfulness,parikh2020totto,ji2023survey,mittal2024towards,filippova2020controlled}. This issue has also been observed in audio foundation models~\cite{sahoo2024comprehensivesurveyhallucinationlarge}. Furthermore, research suggests that pretraining language models for predictive accuracy inherently predisposes them to hallucination, even under ideal conditions where the training data is entirely factual~\cite{kalai2024calibrated}. 

However, few studies explore hallucination evaluation and detection in automatic speech recognition (ASR) systems, with most research focusing on \emph{Whisper}, a semi-supervised model. For instance, \citet{koenecke2024careless} analyzes the Aphasia dataset and reports that while \emph{Whisper}’s overall hallucination rate is 1\%, 40\% of these hallucinations contain violent or harmful content. Similarly, \citet{kim2024automatic} demonstrates that \emph{Whisper} hallucinates at significantly higher rates under low signal-to-noise ratio (SNR) conditions, and observes a 20\% increase in hallucinations at -4 dB and -2 dB SNRs. Prior work by \citet{serai2021hallucinationspeechrecognitionerrors} proposes augmenting models with hallucinated transcripts to improve performance, while \citet{frieske2024hallucinations} introduces a perturbation-based evaluation method using automatic metrics such as word error rate (WER), perplexity, and cosine similarity. \citet{baranski2025investigation} develops a filtered Bag of Hallucinations (BoH) for detection, and reveals that hallucinations in \emph{Whisper} correlate strongly with training data biases (e.g., phrases like ``Thank you for watching'' linked to YouTube content).

Despite these advances, existing studies remain limited in scope, focusing predominantly on semi-supervised models like \emph{Whisper}. This highlights a critical gap: the lack of a comprehensive understanding of hallucinations across the full spectrum of supervision paradigms—from supervised to unsupervised models—and under domain shifts where test data distributions diverge sharply from training environments. Addressing these gaps is essential for developing robust ASR systems that maintain accuracy and faithfulness in diverse real-world applications, a challenge our work directly tackles by evaluating a wide range of models with diverse architectures, sizes, and training paradigms on synthetic and natural shifts.

\section{Methodology}\label{sec:methodology}

We examine transcription quality using the standard metrics WER and CER, and we assess the occurrence of hallucination errors to provide a comprehensive view of model performance. Our testing environment is characterized by both natural and synthetic distribution shifts. Furthermore, we investigate the deterioration of error rates when transitioning from a controlled source domain (LibriSpeech) to various target domains, with a particular focus on both WER and the Hallucination Error Rate (HER). In addition, we introduce noise to the input data and analyze its effect on error rate degradation. This multifaceted strategy offers valuable insights into the challenges encountered by ASR systems in real-world scenarios. We provide additional details about each step subsequently.

\subsection{ASR Evaluation}
We evaluate a broad range of ASR models under a zero-shot setting, using the default decoding parameters for each model. Standard preprocessing steps are applied prior to calculating the metrics, ensuring consistency in evaluation. 

\subsection{Hallucination Evaluation}\label{subsec:hallucination_evaluation}
In addition to conventional transcription errors, we also assess hallucination errors by using an LLM-based pipeline that classifies the errors produced by the ASR model. Specifically, we use \emph{GPT-4o mini} to compare the ground truth transcription with the model outputs and ask the LLM to categorize them into different error types.

We conduct hallucination evaluation at two levels:

\noindent\textbf{Coarse-grained:} The model categorizes the output into one of three classes: \textit{Hallucination Error}, \textit{Non-Hallucination}, or \textit{No Error}. For this evaluation, we provide two examples per category in the prompt.

\noindent\textbf{Fine-grained:} The model is asked to further refine the categorization by identifying specific error types, such as \textit{Hallucination Error}, \textit{Language Error}, \textit{Oscillation Error}, \textit{Phonetic Error}, and \textit{No Error}. In this case, one example per category is provided in the prompt. 

The prompts used for both coarse-grained and fine-grained evaluations are detailed in Appendix~\ref{appsubsec:prompts} (Figure~\ref{fig:coarsegrained_prompt} and Figure~\ref{fig:finegrained_prompt}, respectively). To quantify hallucination occurrences, we introduce the \textit{HER}, defined as the ratio of hallucination errors to the total number of examples in the data.


\begin{figure}[!htp]
    \centering
    \begin{subfigure}{0.24\textwidth}
        \centering
        \includegraphics[width=\linewidth]{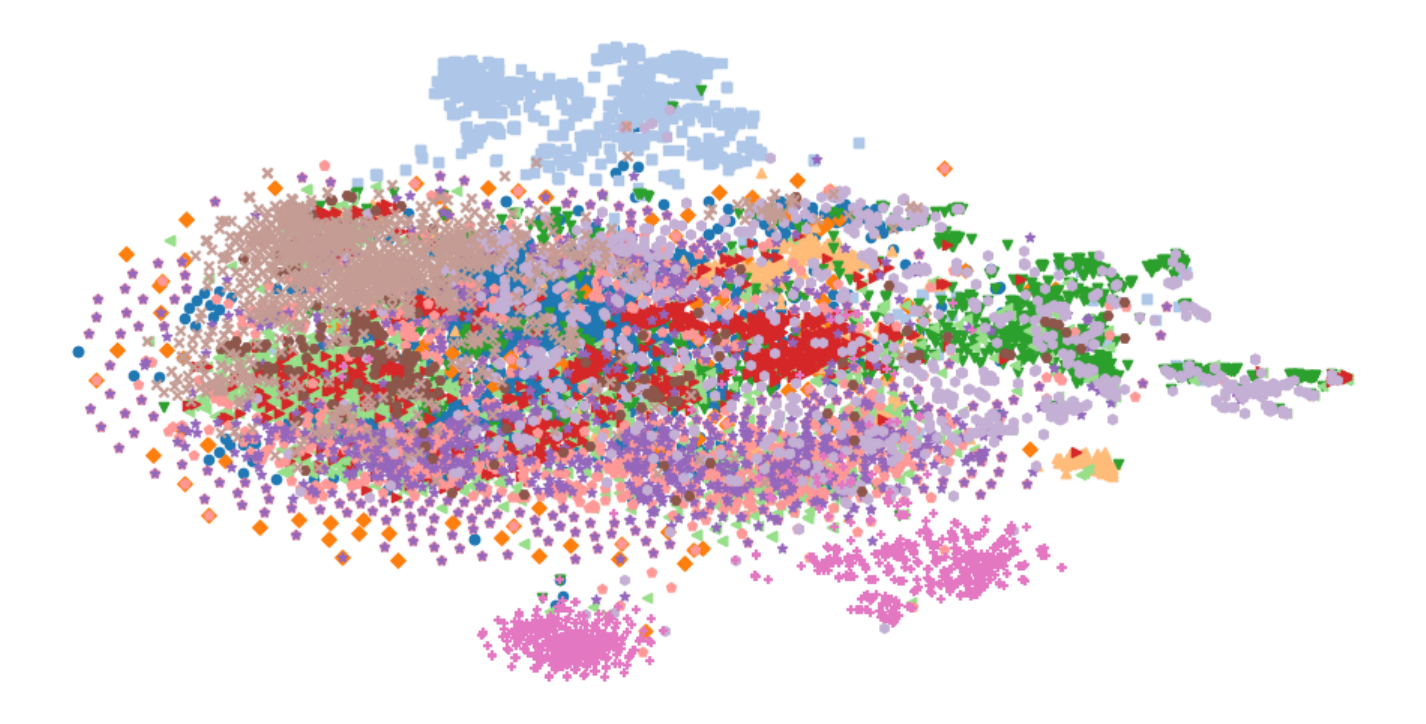}  
        \caption{Speech}
        \label{fig:first_figure}
    \end{subfigure}%
    \hfill
    \begin{subfigure}{0.24\textwidth}
        \centering
        \includegraphics[width=\linewidth]{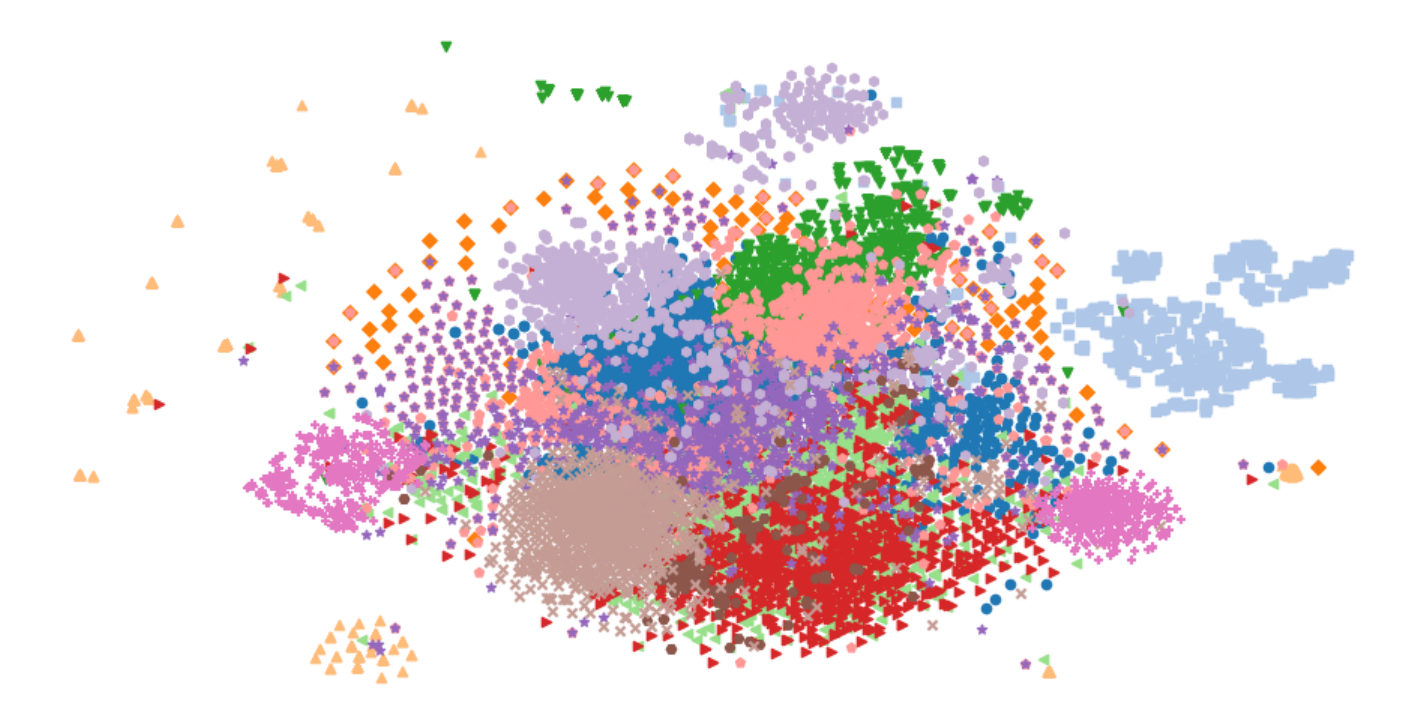}  
        \caption{Text}
        \label{fig:second_figure}
    \end{subfigure}
    \caption{2D t-SNE representation of our evaluation datasets. The figure (a) shows speech representations, while the figure (b) represents text embeddings, both from SONAR. Each distinct evaluation dataset is represented by unique colors and markers, demonstrating the diversity in both the speech and text of our evaluations.}
    \label{fig:speech_text_embeddings}
\end{figure}

\subsection{Distribution Shifts and Quantification}\label{subsec:distribution_shift_and_quantification}
We systematically evaluate ASR models under a variety of testing conditions, ranging from naturally occurring domain variations to scenarios involving both adversarial and non-adversarial perturbations. These conditions are designed to induce either natural or synthetic distribution shifts in the input speech.

\noindent\textbf{Natural Shifts.} These shifts arise from inherent variations in data distributions, such as differences in accents, domain-specific content, background noise, and diverse speaking styles~\cite{liu2021towards}.

\noindent\textbf{Synthetic Shifts.} These shifts are artificially induced, encompassing simulated noise, goal-specific perturbations, and adversarial attacks~\cite{fan2022normalization}. We design a comprehensive testing setup to identify the conditions under which ASR models are prone to hallucinate. Additional details about the datasets used in this study are provided in Section~\ref{subsec:datasets} and summarized in Table~\ref{tab:datasets}.
We measure the extent of domain distribution shifts using the high-order metric Central Moment Discrepancy (CMD)~\cite{zellinger2019centralmomentdiscrepancycmd, kashyap2020domain}, which assesses the discrepancy between two distributions. It is calculated as follows:
\begin{multline}
\text{CMD} = \frac{1}{L} \sum_{l=1}^{L} \left\| \mathbb{E}[h_l^s] - \mathbb{E}[h_l^t] \right\|_2^2,
\end{multline}
where \(L\) represents the number of layers, \(h_l^s\) and \(h_l^t\) denote the hidden representations for the source and target domains, respectively, and \(\mathbb{E}[\cdot]\) signifies the expectation.

\subsection{Error Rate Degradation}
Error Rate Degradation quantifies the decline in ASR performance when transitioning from a source domain (LibriSpeech) to a target domain, with degradation measured in two aspects: transcription errors and hallucination errors.

\noindent\textbf{Word Error Rate Degradation (WERD).} WERD is defined as the difference in WER between the target and source domains:
\begin{equation}
\text{WERD} = \text{WER}_{\text{target}} - \text{WER}_{\text{source}},
\label{eq:2}
\end{equation}
where \(\text{WER}_{\text{source}}\) and \(\text{WER}_{\text{target}}\) denote the WER for the source and target domains, respectively.\\

\noindent\textbf{Hallucination Error Rate Degradation (HERD).} HERD captures the increase in hallucination errors when moving from the source to the target domain:
\begin{equation}
\text{HERD} = \text{HER}_{\text{target}} - \text{HER}_{\text{source}},
\label{eq:3}
\end{equation}
where \(\text{HER}_{\text{source}}\) and \(\text{HER}_{\text{target}}\) represent the hallucination error rates in the source and target domains, respectively.

Furthermore, the relationship between CMD and degradation rates is analyzed and visualized (see Figure \ref{fig:werd_herd_with_shift_all}) to understand how domain variations correlate with both transcription and hallucination errors.

\section{Experiments}\label{sec:experiments}
In this section, we present the experimental details of our work. We examine the models' tendencies to produce hallucinated outputs, using LLM-based evaluation as described in~\ref{sec:methodology}. The results provide insights into the reliability of different ASR systems across various real-world and adversarial conditions.

\subsection{Datasets}\label{subsec:datasets}
In our experiments, we use a diverse set of datasets representing various domains and testing conditions to evaluate the ASR systems under scenarios that differ from training data. To achieve this, we choose datasets from domains with a high likelihood of being unseen during training, ensuring a natural distributional shift. Additionally, we include datasets with synthetic perturbations, such as adversarial attacks, and common augmentation techniques like Gaussian noise addition, pitch shifting, and time stretching, to assess the model's robustness under synthetic shift.

\paragraph{Domain Specific Datasets.} To evaluate model performance under real-world conditions, we leverage datasets from diverse domains, ensuring a comprehensive assessment across various settings. These include legal proceedings: \emph{Supreme-Court-Speech}\footnote{\url{https://huggingface.co/datasets/janaab/supreme-court-speech}}, medical dialogues: \emph{Primock57} \cite{papadopoulos-korfiatis-etal-2022-primock57}, meeting conversations: \emph{AMI} \cite{ami_corpus}, aviation communications: \emph{ATCOsim} \cite{hofbauer-etal-2008-atcosim}, conversational speech: \emph{SLUE-VoxCeleb} \cite{shon2022slue}, home environments: \emph{BERSt}\footnote{\url{https://huggingface.co/datasets/macabdul9/BERSt}}, and general speech corpora: \emph{LibriSpeech} \cite{panayotov2015librispeech}, \emph{GLOBE} \cite{wang2024globe}, and \emph{SPGISpeech} \cite{kensho2021spgispeech}, including noisy conditions (\emph{LibriSpeech$\_$test$\_$noise} \cite{panayotov2015librispeech}). These datasets span a wide range of accents, recording conditions, and environments—from high-quality audiobooks to teleconferences and real-time simulations. This diversity ensures a robust evaluation of model performance across realistic and challenging scenarios, addressing the limitations of single-domain evaluations.

\paragraph{Perturbed Datasets.} To simulate challenging acoustic conditions and evaluate WER and HER under adversarial scenarios, we apply various types of synthetic perturbations to speech inputs. These include an adversarial dataset featuring modified utterances with adversarial noise at varying radii (0.04 and 0.015) and Room Impulse Response (RIR) noise, primarily aimed at adversarially attacking the ASR models~\cite{Olivier2022RI}. Additionally, we evaluate model robustness under challenging conditions by applying a range of general audio perturbations—including noise addition, time stretching, pitch shift, cross-lingual noise, distortion, echo, pub noise, and reverberation—to 1,000 randomly sampled audio clips from a mixture of domain-specific datasets. These perturbations are commonly used as augmentation techniques to simulate real-world variability and stress-test the models. We compare the performance on perturbed speech with that of non-perturbed speech to quantify the impact of these distortions.
Comprehensive information about the datasets used in this study can be found in Appendix~\ref{appsubsubsec:datasets} (Table~\ref{tab:datasets}).

\subsection{Models}\label{subsec:models}

To comprehensively evaluate hallucination patterns in ASR systems, we select models that span diverse architectures, sizes, and training paradigms. This diversity enables systematic analysis of how these factors influence hallucination susceptibility. Specifically, we include:  
\begin{itemize}  
    \item \emph{Encoder-only} models: HuBERT~\cite{hsu2021hubert} and Wav2Vec2~\cite{baevski2020wav2vec}, which leverage self-supervised training to learn robust speech representations.  
    \item \emph{Encoder-decoder} models: Whisper~\cite{radford2022robustspeechrecognitionlargescale} (10 variants), DistilWhisper~\cite{gandhi2023distilwhisper} (4 variants), and SeamlessM4T~\cite{communication2023seamlessm4tmassivelymultilingual} (2 variants) are designed for multilingual transcription, translation, and speech-to-text tasks. We also include Qwen2Audio~\cite{chu2024qwen2} and SpeechLLM~\cite{Rajaa_SpeechLLM_Multi-Modal_LLM}, which are optimized for text generation and audio-language alignment. Additionally, we incorporate Canary~\cite{Harper_NeMo_a_toolkit}\footnote{\url{https://huggingface.co/nvidia/canary-1b}}, which uses FastConformer~\cite{rekesh2023fastconformerlinearlyscalable} as its encoder and a transformer decoder.

\end{itemize}

In addition to these, we also include open-weight transducer-based models. Specifically, we add a model trained with transducer decoder loss (\emph{parakeet-tdt-1.1b}) and a token-and-duration transducer~\cite{xu2023efficientsequencetransductionjointly} (\emph{parakeet-tdt-1.1b}), both from the NVIDIA-NeMo ASR repository~\cite{Harper_NeMo_a_toolkit}. In summary, the selected models vary in size (39M to 7B parameters), depth (4 to 32 layers), and training paradigms (supervised, self-supervised, and semi-supervised). Full specifications, including architectural details, training data, and hyperparameters, are provided in Appendix~\ref{appsubsecmodels} (Table~\ref{tab:models}).

\subsection{Experimental Setup}
We utilize models and datasets sourced from Huggingface~\footnote{\url{https://huggingface.co/models,https://huggingface.co/datasets}}. All audio data is resampled to match the sampling rate required by the respective models. For each dataset, we randomly sample 1,000 examples from the \textit{test} split to ensure a manageable and consistent experimental setup. Unless otherwise specified, we use the default decoding parameters for ASR evaluation. To analyze the data, we compute SONAR embeddings~\cite{duquenne2023sonar} for both speech and text. Additionally, we employ CMD based on prior work~\cite{kashyap2020domain} to quantify domain shifts. All experiments are conducted on a single A100/H100 GPU. Prior to calculating WER and generating embeddings, we apply a basic English text normalizer~\footnote{\url{https://github.com/huggingface/transformers/blob/main/src/transformers/models/whisper/english_normalizer.py}} to ensure consistency in text preprocessing. For LLM evaluation, we perform greedy search decoding to ensure reproducible outputs.

\subsection{Human Evaluation}
We construct a human evaluation dataset by aggregating outputs from multiple models, filtering samples with \texttt{WER} > 60 to focus on significant deviations. Hypotheses and references are constrained to 1-100 words for balance. To simulate synthetic hallucinations, we shuffle 50 hypotheses, introducing artificial errors~\cite{stiff2019improving}. The final dataset includes 500 samples, each reviewed by two independent annotators from a pool of 20. This framework ensures robust evaluation and reliable analysis of hallucination patterns across models.

\section{Results}\label{sec:results}
In our experiments, we use an LLM-based pipeline to classify ASR errors across various evaluation setups, validating it against human evaluation and heuristic baselines. We then explore the effects of natural and synthetic distribution shifts on error metrics, specifically examining how domain variations and input perturbations impact word and hallucination error rates. Additionally, we analyze the influence of model architecture and scale through a comparison of Whisper variants and other architectures. This approach provides insights into the complex interactions between error types, data conditions, and model characteristics. In this section, we present our findings.

\subsection{Hallucination Error Detection}

We assess the ASR outputs of large language models by classifying them into various error categories. Specifically, we use \emph{GPT-4o mini} to identify the types of errors in ASR outputs. The evaluation process includes both coarse-grained and fine-grained error classifications, as detailed in Section~\ref{subsec:hallucination_evaluation}. The prompts used for both evaluations are shown in Figures~\ref{fig:coarsegrained_prompt} and~\ref{fig:finegrained_prompt}.



\paragraph{Verbalized Confidence Scores}

Accurate error classification alone does not always provide insight into the model’s own uncertainty or reliability. To address this, we also assess the self-reported confidence of the LLM in its classifications. Inspired by~\citet{yang2024verbalizedconfidencescoresllms} and other similar works~\cite{tian2023justaskcalibrationstrategies, xiong2024llmsexpressuncertaintyempirical}, we prompt GPT-4o-mini to provide explicit confidence scores (on a scale of 1–10) for each classification decision. This approach allows us to understand not just what the model predicts, but how confident it is in those predictions, offering an additional dimension for evaluating model performance and robustness. In this experiment, we take 500 examples, and after each classification, we simply ask the model in a multiturn conversation: ``\textit{How confident are you about the classification on a scale of 1–10? Confidence:}''. We report the resulting confidence scores in Tables~\ref{tab:confidence_finegrained} and~\ref{tab:confidence_coarsegrained} for fine-grained and coarse-grained evaluations, respectively.

The results show that GPT-4o-mini’s self-assessed confidence remains consistently high across most categories. In the fine-grained evaluation, the ``Hallucination Error'' (HE) and ``Oscillation Error'' (OE) classes achieve nearly perfect confidence scores, with median and 25\%ile values of 10.0. The ``No Error'' (NE) predictions also exhibit strong confidence, with an overall average of 8.81 across fine-grained classes, highlighting robust certainty in identifying transcription errors.
\begin{table}[ht!]
\renewcommand{\arraystretch}{1}
\centering
\resizebox{\linewidth}{!}{%
\begin{tabular}{lcccccc}
\toprule
\textbf{Metric} & \textbf{HE} & \textbf{LE} & \textbf{NE} & \textbf{OE} & \textbf{PE} & \textbf{Overall} \\
\midrule
Mean     & 9.56 & 7.33 & 9.77 & 10.0 & 7.37 & 8.81 \\
Median   & 10.0 & 7.0  & 10.0 & 10.0 & 7.0  & 8.80 \\
1\%ile   & 8.00 & 7.00 & 7.21 & 10.0 & 7.00 & 7.84 \\
5\%ile   & 8.0  & 7.0  & 8.1  & 10.0 & 7.0  & 8.02 \\
10\%ile  & 8.0  & 7.0  & 10.0 & 10.0 & 7.0  & 8.40 \\
25\%ile  & 9.0  & 7.0  & 10.0 & 10.0 & 7.0  & 8.60 \\
\bottomrule
\end{tabular}
}
\caption{Self-verbalized confidence scores (1–10) of the LLM classifier for each fine-grained error class—Hallucination Error (HE), Language Error (LE), No Error (NE), Oscillation Error (OE), and Phonetic Error (PE)—along with overall average scores.}

\label{tab:confidence_finegrained}
\end{table}

In contrast, the ``Language Error'' (LE) and ``Phonetic Error'' (PE) classes show slightly lower confidence, with mean values around 7.3 and narrower interquartile ranges, reflecting more nuanced or ambiguous cases.

\begin{table}[ht!]
\centering
\begin{tabular}{lcccc}
\toprule
\textbf{Metric} & \textbf{HE} & \textbf{NE} & \textbf{NHE} & \textbf{Overall} \\
\midrule
Mean     & 9.77 & 9.20 & 7.09 & 8.69 \\
Median   & 10.0 & 10.0 & 7.0  & 9.00 \\
1\%ile   & 8.0  & 7.0  & 1.0  & 5.33 \\
5\%ile   & 8.0  & 7.95 & 7.00 & 7.65 \\
10\%ile  & 9.0  & 8.0  & 7.0  & 8.00 \\
25\%ile  & 10.0 & 8.0  & 7.0  & 8.33 \\
\bottomrule
\end{tabular}
\caption{Self-verbalized confidence scores (1–10) of the LLM classifier for each coarse-grained error class—Hallucination Error (HE), No Error (NE), and Non-Hallucination Error (NHE)—along with overall average scores.}

\label{tab:confidence_coarsegrained}
\end{table}

The coarse-grained evaluation mirrors these trends, with ``Hallucination Error'' and ``No Error'' classes maintaining high confidence (mean $\geq$ 9.2), while the ``Non-Hallucination Error'' (NHE) class has a lower mean of 7.09 and a notably low 1\%ile of 1.0, indicating occasional uncertainty in ambiguous scenarios. Despite this, the overall average for the coarse-grained setting is 8.69, confirming the reliability of these scores.

Overall, these findings demonstrate that GPT-4o-mini not only provides reliable classification outputs but also accurately calibrates its confidence based on the error type. The high overall averages further validate the robustness of the LLM-based evaluation framework in discerning and categorizing transcription errors.


\paragraph{Coarse-grained and Fine-grained Evaluation.} Figure~\ref{fig:pie_finegrained} in the appendix illustrates the error distributions across coarse-grained and fine-grained hallucination categories. Our results demonstrate strong alignment between both levels, indicating consistent classification of hallucinations. Among non-hallucination errors, phonetic errors dominate across most datasets (see Appendix Table~\ref{tab:non_hallucination_error_analysis}). However, in \emph{Primock57}, language errors prevail, likely due to its specialized medical terminology. This aligns with~\cite{ferrando2024know}, who emphasize language models' struggles with domain-specific named entities. This is also reflected in the second example provided in Table~\ref{tab:examples}.

\paragraph{Agreement with Human Evaluation and Heuristic Baseline.} To validate our approach, we compare \emph{model-to-human} and \emph{human-to-human} agreement scores using a coarse-grained prompt. Our results demonstrate strong human-to-human raw agreement (0.71), indicating consistency. Additionally, we observe good agreement (0.6) between human annotations and \emph{GPT-4o-mini}'s coarse-grained output, suggesting that the model aligns reasonably well with human judgments.


\begin{table}[h]
\centering
\renewcommand{\arraystretch}{1.2}
\begin{tabular}{lc}
\toprule
\textbf{Evaluation Pair} & \textbf{Agreement Score ($\uparrow$)} \\
\midrule
Human - Human & 0.71 $\pm$ 0.01 \\
Human - Heuristic & 0.00 $\pm$ 0.00 \\
Human - GPT & 0.60 $\pm$ 0.01 \\
Human - Gemini & 0.59 $\pm$ 0.01 \\
GPT - Heuristic & 0.10 $\pm$ 0.00 \\
GPT - Gemini & 0.78 $\pm$ 0.01 \\
Heuristic - Gemini & 0.14 $\pm$ 0.01 \\
\bottomrule
\end{tabular}
\caption{Agreement scores ($\uparrow$) between different hallucination evaluation methods. Here, \textit{GPT-4o-mini} is referred to as GPT, and \textit{Gemini-2.0-flash-001} is referred to as Gemini. All evaluations are coarse-grained.}
\label{tab:agreement_scores}
\end{table}

We further evaluate the agreement between human and model classifications against a heuristic baseline proposed by~\cite{frieske2024hallucinations}. Their method is based on a cosine similarity threshold of 0.2, alongside an \emph{WER} threshold of 30 and a \emph{Flan-T5}~\cite{chung2024scaling} perplexity threshold of 200. However, as shown in Table~\ref{tab:agreement_scores}, this heuristic achieves significantly lower agreement scores: 0.1 with \emph{GPT-4o mini} and 0.14 with \emph{Gemini-2.0-flash-001}. These results highlight the limitations of purely heuristic-based approaches compared to our method, which better captures the more fine-grained aspects like hallucination.

\begin{table*}[ht!]
\resizebox{\textwidth}{!}{%
\begin{tabular}{lccccccccccccc}
\toprule
\multicolumn{1}{l}{\multirow{1}{*}{Model}} & 
\multicolumn{1}{c}{SPGI} & 
\multicolumn{1}{c}{BERSt} & 
\multicolumn{1}{c}{ATCOsim} & 
\multicolumn{1}{c}{ADV} & 
\multicolumn{1}{c}{AMI} & 
\multicolumn{1}{c}{SLU} & 
\multicolumn{1}{c}{SNIPS} & 
\multicolumn{1}{c}{SC} & 
\multicolumn{1}{c}{GLOBE} & 
\multicolumn{1}{c}{SALT} & 
\multicolumn{1}{c}{LS\_Noise} & 
\multicolumn{1}{c}{LS} & 
\multicolumn{1}{c}{Primock57} \\
\midrule
whisper-large-v3 & 3.4/0.4 & 32.4/15.8 & 65.3/17.6 & 33.3/49.8& 23.4/10.1 & 15.5/7.9 & 8.2/\colorbox[HTML]{8fbc8f}{0.9} & 18.8/15.4 & 3.4/2.5 & 3.0/\colorbox[HTML]{8fbc8f}{1.0} & 2.6/\colorbox[HTML]{8fbc8f}{0.4} & 2.2/0.3 & 19.2/\colorbox[HTML]{8fbc8f}{4.5} \\ 
wav2vec2-large-xlsr-53-english & 19.6/1.0 & 64.0/18.6 & 63.0/19.6 & 100.1/96.5 & 53.0/22.7 & 43.4/6.2 & 12.4/1.2 & 32.6/14.8 & 27.0/10.3 & 17.0/2.1 & 9.0/0.6 & 6.5/\colorbox[HTML]{8fbc8f}{0.0} & 47.9/13.7 \\ 
hf-seamless-m4t-large & 14.7/3.6 & 58.9/34.6 & 76.6/62.3 & 61.2/76.9 & 63.7/46.2 & 44.0/23.5 & 7.4/3.2 & 34.2/30.0 & 19.2/21.4 & 4.2/1.6 & 6.5/3.2 & 3.4/0.5 & 44.5/27.4 \\ 
speechllm-1.5B & 11.5/4.8 & 68.5/38.5 & 121.4/57.4 & 95.3/94.5 & 127.3/54.4 & 83.8/16.5 & 10.8/2.9 & 41.3/38.4 & 27.9/28.7 & 9.5/4.2 & 10.9/5.2 & 11.4/4.2 & 41.7/17.1 \\ 
whisper-medium & 3.7/\colorbox[HTML]{8fbc8f}{0.2} & 34.5/16.2 & 65.6/18.2 & 42.9/63.1 & 23.2/12.2 & 17.4/8.7 & 8.6/1.4 & 18.7/14.4 & 5.3/3.1 & 5.0/3.7 & 3.3/0.9 & 3.1/0.4 & 20.6/6.2 \\ 
distil-large-v2 & 4.1/1.0 & 38.0/16.0 & 69.5/29.6 & 45.6/64.3 & 22.1/11.2 & 16.0/7.2 & 9.2/1.3 & 18.9/14.3 & 6.7/3.2 & 5.2/1.0 & 3.6/0.7 & 3.4/0.5 & 19.2/5.2 \\ 
hubert-large-ls960-ft & 12.4/1.4 & 58.5/\colorbox[HTML]{8fbc8f}{14.3} & 50.0/\colorbox[HTML]{8fbc8f}{11.0} & 109.8/100.0 & 44.4/29.5 & 21.3/\colorbox[HTML]{8fbc8f}{2.4} & 12.6/1.2 & 30.2/20.4 & 23.4/7.1 & 18.7/3.7 & 3.6/1.3 & 2.2/0.1 & 32.2/12.0 \\
distil-medium.en & 4.6/0.6 & 39.3/17.5 & 71.3/34.0 & 45.8/63.9 & 23.6/8.5 & 15.6/7.1 & 9.7/1.4 & 20.0/14.5 & 8.5/3.4 & 7.4/3.7 & 4.3/1.7 & 4.2/0.9 & 21.0/5.7 \\ 
distil-small.en & 4.6/0.6 & 46.8/19.4 & 77.0/41.6 & 54.3/73.3 & 24.2/\colorbox[HTML]{8fbc8f}{8.0} & 15.4/7.8 & 11.3/2.3 & 21.5/18.2 & 11.7/8.1 & 9.0/4.7 & 4.1/0.9 & 4.0/0.6 & 21.4/6.4 \\ 
whisper-medium.en & 4.3/1.0 & 34.2/18.8 & 66.6/22.8 & 43.3/60.4 & 23.0/11.3 & 19.4/9.5 & 8.4/1.5 & 21.3/15.4 & 4.8/2.3 & 5.7/3.7 & 3.5/0.9 & 3.1/0.4 & 20.6/5.7 \\
whisper-small.en & 4.1/0.9 & 38.7/19.7 & 68.8/29.2 & 50.9/72.9 & 24.5/13.7 & 20.8/10.7 & 9.4/1.2 & 20.9/16.4 & 9.6/6.4 & 7.2/5.8 & 3.7/0.9 & 3.6/0.3 & 21.5/6.7 \\ 
hf-seamless-m4t-medium & 13.2/4.4 & 57.9/32.1 & 52.7/50.7 & 51.4/67.1 & 57.0/43.7 & 50.3/24.9 & 8.8/2.1 & 36.0/31.3 & 15.9/16.1 & 6.4/2.1 & 8.9/2.3 & 3.7/0.5 & 46.1/26.3 \\ 
whisper-tiny & 8.8/3.6 & 122.1/41.9 & 110.3/63.9 & 88.0/90.2 & 40.3/26.3 & 22.5/12.3 & 15.6/5.5 & 38.6/31.6 & 54.7/50.6 & 20.0/13.6 & 10.8/7.1 & 7.6/1.7 & 32.8/16.8 \\ 
whisper-large & 3.7/0.6 & 48.1/15.8 & 65.7/18.5 & 37.2/55.7 & 22.6/12.9 & 18.1/9.1 & 8.5/1.0 & 18.6/14.8 & 4.2/2.5 & 4.0/2.6 & 3.1/0.7 & 3.0/0.1 & 20.0/5.7 \\ 
whisper-large-v2 & 4.3/0.8 & 34.1/17.9 & 67.1/19.6 & 38.9/57.6 & 24.1/15.8 & 18.2/11.2 & 8.4/1.1 & 23.6/15.5 & 4.4/3.6 & 3.2/\colorbox[HTML]{8fbc8f}{1.0} & 2.7/0.6 & 3.0/0.3 & 20.0/6.9 \\ 
whisper-large-v3-turbo & 3.4/0.4 & 31.7/14.7 & 66.2/18.7 & 34.5/49.8 & 23.8/10.0 & 15.7/7.4 & 7.8/\colorbox[HTML]{8fbc8f}{0.9} & 18.5/\colorbox[HTML]{8fbc8f}{13.8} & 3.9/\colorbox[HTML]{8fbc8f}{1.7} & 4.7/2.6 & 2.7/0.5 & 3.3/0.1 & 20.0/6.0 \\ 
Qwen2-Audio-7B & 4.6/2.7 & 36.3/15.6 & 44.8/35.7 & 31.7/\colorbox[HTML]{8fbc8f}{46.7 }& 35.7/14.9 & 47.4/32.1 & 5.5/1.3 & 35.3/37.1 & 23.3/7.0 & 5.9/5.8 & 2.3/1.3 & 2.0/0.7 & 25.5/22.8 \\ 
\bottomrule
\end{tabular}%
}
\caption{WER and coarse-grained HER across different models and datasets. The values are presented as WER/HER. The lowest HER for each dataset is highlighted in green. Abbreviations: SPGI (SPGISpeech), ATCOSIM (ATCOsim Corpus), ADV (Adversarial), AMI (AMI Corpus), SLU (SLUE-VoxCeleb), SC (Supreme-Court-Speech), SALT (SALT Multispeaker English), LS\_Noise (LibriSpeech Test Noise), LS (LibriSpeech ASR Test).}
\label{tab:all_results}
\end{table*}

\begin{figure*}[!hbt]
    \centering
    \includegraphics[width=1.0\linewidth]{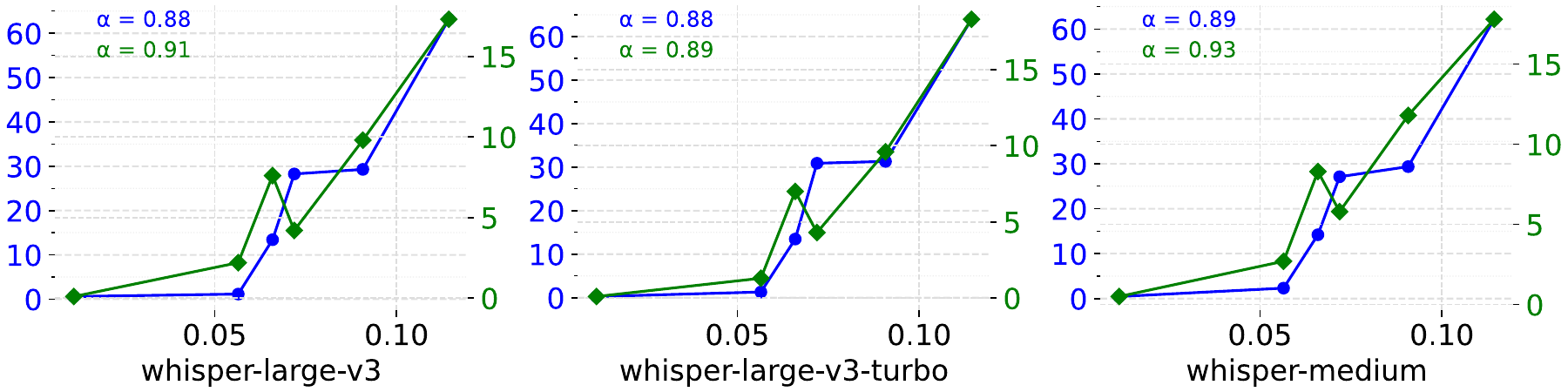}
    \caption{Degradation in word error rate WERD (blue, left y-axis) and hallucinated error rate HERD (green, right y-axis) w.r.t distribution shift (x-axis), measured using Central Moment Discrepancy (CMD) for three different models. The correlation factor, $\alpha$, is represented by the color, which corresponds to the type of error. Each point on the line represents a new target domain.}
    \label{fig:werd_herd_shift}
\end{figure*}


\subsection{Errors Under Distributional Shifts}
We analyze how ASR performance degrades under distributional shifts—when models are exposed to domain conditions different from their training data or domains where they outperform the human baseline (referred to as the source domain)\footnote{Here, ``domain'' is loosely defined; for some models, the source or training data is unknown. In those cases, we treat the source domain as the data on which the error is nearly zero, such as LibriSpeech.}. For natural (domain-specific) shifts, we quantify the distribution shift and measure the resulting degradation in both WER and hallucination errors (HER), with results shown in Figure~\ref{fig:werd_herd_shift} and detailed breakdowns in Appendix~\ref{appsubsec:results}. For synthetic shifts (e.g., adversarial and random perturbations), we directly evaluate model robustness. In what follows, we describe these two shift types—natural and synthetic—separately.

\paragraph{Natural Shift} Given that most ASR models now outperform the human baseline on the LibriSpeech clean test set, we consider LibriSpeech as the source domain. Other domain-specific datasets, such as Primock, SPGISpeech, GLOBE, and AMI, are therefore treated as the target domain. We compute the distribution shift as detailed in Section~\ref{subsec:distribution_shift_and_quantification}. We then measure the change (degradation) using Equation~\ref{eq:2} and \ref{eq:3}.
 
The $\alpha$ is the correlation coefficient between error rate degradation and distribution shift. Figure~\ref{fig:werd_herd_shift} shows that both WER and HER degrade as we move from the source domain to different target domains, with considerable distribution shifts across various \emph{whisper} models. This degradation exhibits a nearly linear positive correlation with the domain shift.

Notably, the HER demonstrates a slightly stronger correlation with the shift compared to WER. This trend is consistent across all models, as illustrated in Appendix~\ref{appsubsec:results} Figure~\ref{fig:werd_herd_with_shift_all}. The \emph{ATCOsim} dataset is an outlier, with artificially high \texttt{WER} due to its numerical content. For example, models generating digits (e.g., "23") instead of spoken forms ("two three") are heavily penalized, inflating \texttt{WER} without accurately reflecting transcription quality.

\paragraph{Synthetic Shift}
Under synthetic shift, we experiment with two configurations: (a) adversarial perturbations and (b) common perturbations. Our experiments reveal that adversarial attacks cause the most significant degradation in HER, with adversarial datasets showing the highest HER values across all models, exceeding the degradation observed under natural shift baselines (see Table~\ref{tab:all_results}). In contrast, random perturbations (such as white noise, pitch shifts, and time stretching) result in more moderate impacts. Notably, self- and semi-supervised models like~\emph{whisper} and~\emph{seamless} demonstrate consistent vulnerability to structured perturbations. For instance, pitch shifts and time stretching lead to a substantial increase of approximately (242\%) in both WER and HER across these models, while white noise causes smaller degradations of approximately (142\%). Interestingly, the supervised~\emph{wav2vec2} model exhibits non-uniform behavior, where the impact on WER and HER is similar across all perturbations. Furthermore, it is noteworthy that HER increased by 50\% in the~\emph{wav2vec2} model, which is considerably less than the sharp increase observed in~\emph{whisper} and~\emph{seamless} models, highlighting a notable contrast in the robustness of these models to random synthetic shifts against more targeted shifts. We provide results for additional perturbations in Appendix~\ref{appsubsubsec:synthetic_shift} Table~\ref{tab:additional_perturbations}.

\begin{figure}[!hbt]
    \centering
    \includegraphics[width=1\linewidth]{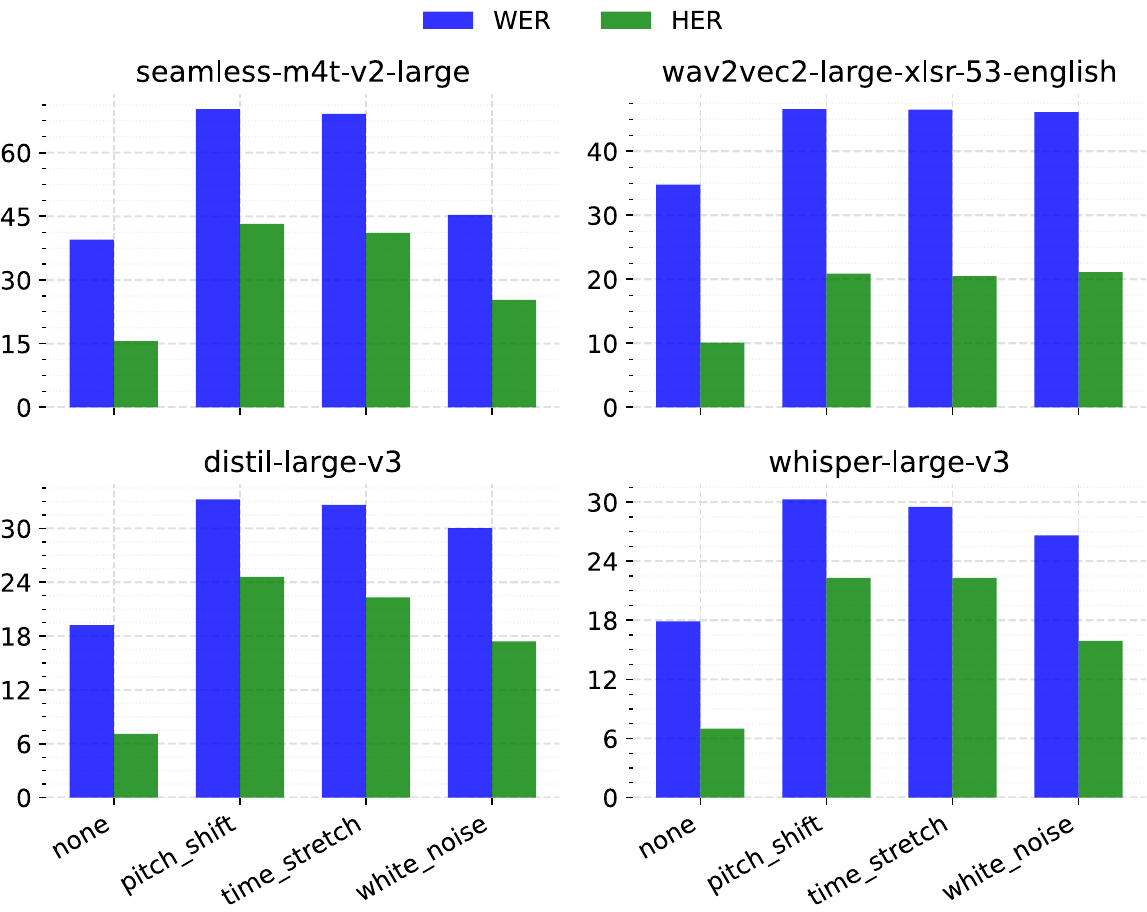}
    \caption{HER and WER for various perturbations across four models.}
    \label{fig:perturb}
\end{figure}


\subsection{Impact of Architecture and Scale}
In this section, we analyze how model architecture and scale influence hallucination and transcription errors. We specifically examine how different model types—encoder-only, encoder-decoder, and transducer-based—affect hallucination error rates, as well as how scaling within a model family (such as \emph{whisper}) impacts overall performance.

\paragraph{Model Type} We hypothesize that model architecture plays a critical role in influencing error rates, with these effects further modulated by the scale and diversity of the training data. As shown in Appendix Table~\ref{tab:results_ratio}, our findings indicate that encoder-only models \emph{Wav2vec2-large-xlsr-53-english} and \emph{Hubert-large-ls960-ft} exhibit the lowest \texttt{HER/WER} ratios across multiple datasets, particularly when compared to models like \emph{Qwen2-Audio-7B} and \emph{hf-seamless-m4t-large}. This suggests that encoder-only models are less prone to hallucinations relative to their overall errors.


\paragraph{Model Size} In our experiments, we evaluate models of varying sizes where training data is fixed across different sizes (ie: \emph{whisper}). 
\begin{figure}[!hbt]
    \centering
    \includegraphics[width=1.0\linewidth]{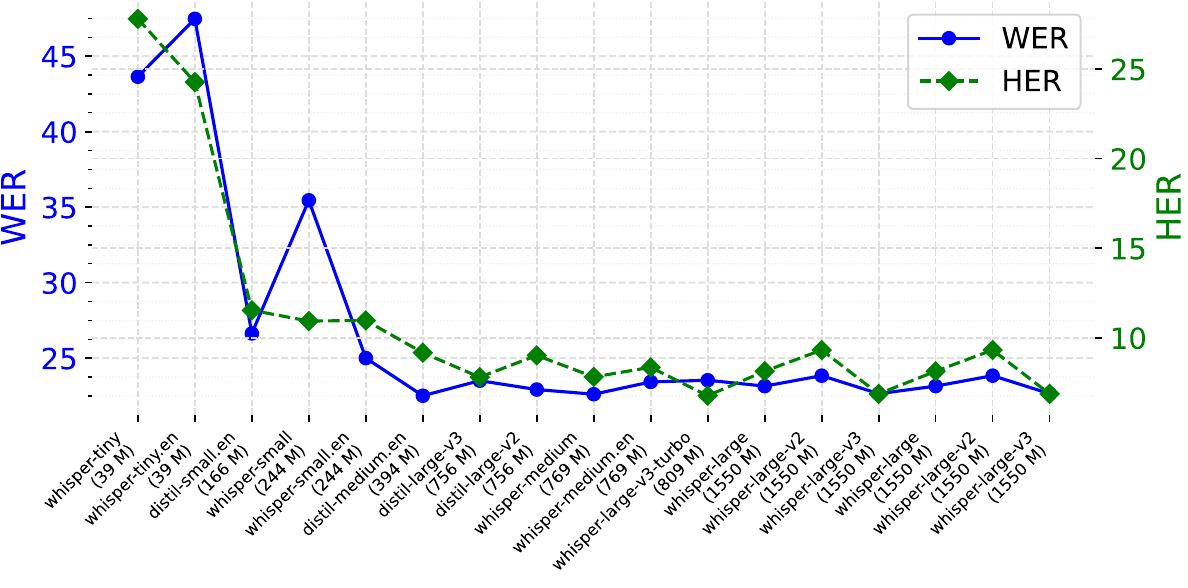}
    \caption{HER (green, right y-axis) and WER (blue, left y-axis) averaged across all datasets. X-axis denotes models with different size in increasing order.}
    \label{fig:werd_herd_whisper}
\end{figure}


More specifically, we select \nummodels~ models from the \emph{whisper} family, ranging from the smallest \emph{whisper-tiny} (39M parameters) to the largest \emph{whisper-large-v3} (1.5B parameters). We then calculate both WER and HER for each of these models, following the methodology outlined in Section~\ref{sec:methodology}. Our findings show that for smaller models, such as \emph{whisper-tiny} and \emph{whisper-small}, there is a significant increase in both WER and HER. While larger models such as \emph{whisper-medium} and \emph{whisper-large} show substantial improvements, as shown in Figure~\ref{fig:werd_herd_whisper}. However, this reduction is not linear. After a certain point, performance improvements in both metrics become less pronounced.
This non-monotonic behavior is particularly evident when comparing models in the mid-range of parameter sizes, such as \emph{whisper-medium} and \emph{whisper-large-v3-turbo}, where the difference in performance becomes marginal despite the difference in model size. In conclusion, while larger models generally result in lower WER and HER, the benefits of scaling up model size diminish beyond a certain point, at least when it comes to more nuanced error types.

\section{Conclusion}\label{sec:conclusion}

In our work, we introduce the Hallucination Error Rate (HER) as a crucial complement to traditional ASR evaluation metrics like WER, especially in high-risk applications where model reliability is critical. By developing a robust LLM-based hallucination detection framework, we present a comprehensive evaluation of ASR models across both synthetic and natural distribution shifts, highlighting the specific challenges ASR systems face under real-world conditions. Our findings emphasize the importance of incorporating HER into standard ASR evaluation practices, particularly for applications in safety-critical domains such as healthcare, legal, and aviation. Through detailed analysis, we show that traditional metrics like WER can mask significant hallucinations, emphasizing the need for more holistic evaluation methods. Our work lays the ground for future work in ASR model reliability, aiming to ensure that ASR systems not only produce accurate transcriptions but also avoid generating misleading, harmful, and unfaithful to input speech transcriptions. In future work, we plan to expand our evaluation to cover additional evaluation setups, ensuring a comprehensive set of assessments. Additionally, we aim to explore mitigation strategies, which are critical for enhancing the reliability of ASR systems.

\section{Limitations}\label{sec:limitations}
We explore the hallucination phenomenon in ASR systems focusing on the potential causes such as distribution shift, model types, and model size and architectures. While our work offers valuable insights into model behavior across different conditions, there are several limitations to consider.

\noindent\textbf{Evaluation Datasets.} 
In our study, we evaluate ASR models across multiple domains, including legal, medical, and conversational speech, ensuring a broad range of datasets not seen during training. However, it is possible that some of the datasets we treat as target domains may have been inadvertently
 exposed to the models during training. Furthermore, the lack of access to a diverse variety of domain-specific datasets limits our understanding of how these models will perform in more diverse or previously unseen domains, particularly those with limited or noisy data. While our focus on domain shifts is an important step, further research is needed to assess model performance in even more varied and challenging real-world environments.

\noindent\textbf{Synthetic Noise and Perturbations.}
Our experiments also include synthetic noise and perturbations to evaluate model robustness. While this approach helps simulate real-world challenges, it does not capture all possible distortions that may occur in uncontrolled environments. Adversarial noise, pitch shifts, and time stretching are some of the perturbations we consider, but other potential real-world disruptions, such as cross-lingual noise or complex acoustic reverberations, are not fully explored.

\noindent\textbf{Hallucination Detection.} 
The detection of hallucinations in ASR systems, as measured by the Hallucination Error Rate (HER), is a key contribution of our study. However, our reliance on LLM-based classifiers introduces potential biases and variability. While we observe strong alignment with human judgments, the accuracy of these evaluations may be influenced by subjective interpretation, especially in edge cases where the boundaries between errors are unclear. Additionally, while LLM-based methods present a novel approach, their performance in low-resource settings or with models trained on limited data has not been fully explored. Furthermore, the use of proprietary models, such as those from OpenAI via API, introduces additional costs, which could limit the scalability of this approach.

\section{Ethics Statement}\label{sec:ethics_statement}

\noindent\textbf{Data Collection and Release.} 
For this study, we rely on publicly available datasets from diverse domains to evaluate hallucinations in ASR systems. We ensure that the data used in our research is appropriately sourced, maintaining respect for copyright, license, and privacy regulations. Furthermore, we emphasize that the use of these datasets is strictly for academic purposes, aligned with the principles of fair use. 

\noindent\textbf{Intended Use.} 
Our work aims to enhance the robustness of ASR systems, especially in high-stake domains where errors can have significant consequences. We believe our findings will encourage further research in hallucination detection, with particular attention to models' performance in low-resource and critical domains such as healthcare and law. By introducing the Hallucination Error Rate (HER) as a complementary metric to traditional evaluation methods, we hope to inspire the development of more reliable and transparent ASR systems.

\noindent \textbf{Potential Misuse and Bias.}
While our work provides valuable insights about hallucinations in ASR systems, we acknowledge that they could be misused if deployed in inappropriate contexts. Since these models are trained on a variety of data sources, there is the potential for them to generate biased or harmful content, especially if the training data contains any inherent biases. Moreover, hallucinations in ASR outputs, if undetected, can lead to severe consequences in critical applications such as legal, medical, and financial settings. We recommend careful deployment of these models, ensuring that they undergo rigorous bias mitigation and hallucination detection processes before being used in such domains.

\bibliography{custom}

\newpage

\appendix
\section{Appendix}

\subsection{Hallucination}
Table~\ref{tab:hallucination_examples} illustrates the key differences between phonetic errors and hallucinations in ASR outputs. In the given example, the reference sentence "It's hard to recognize speech" is misrecognized phonetically as "It's hard to wreck a nice beach", demonstrating a typical pronunciation-based error where acoustically similar words are confused. In contrast, the hallucination example shows the model generating "Now move to me" instead of the reference "There are more coming", introducing semantically unrelated content not supported by the input speech. This highlights that phonetic errors stem from acoustic-perceptual confusions, while hallucinations involve the model producing arbitrary, contextually inconsistent text. The distinction is crucial for analyzing ASR failures, as each type requires different mitigation strategies—improving acoustic modeling for phonetic errors versus constraining language generation for hallucinations.

\subsection{Experiments}\label{appsubsec:experiments}
In this section, we provide more details about our experimental setup.
\subsubsection{Datasets}\label{appsubsubsec:datasets}
We provide a brief description of the datasets we use in our experiment in Table~\ref{tab:datasets}.

\subsubsection{Models}\label{appsubsecmodels}
We provide a brief description of the models we use in our experiment in Table~\ref{tab:models}.

\subsection{Prompts}\label{appsubsec:prompts}
We show the prompts for coarse-grained and fine-grained error classification in Figures~\ref{fig:coarsegrained_prompt} and~\ref{fig:finegrained_prompt}, respectively.

\subsection{Results}\label{appsubsec:results}

\begin{figure*}
\centering
\includegraphics[width=1.0\linewidth]{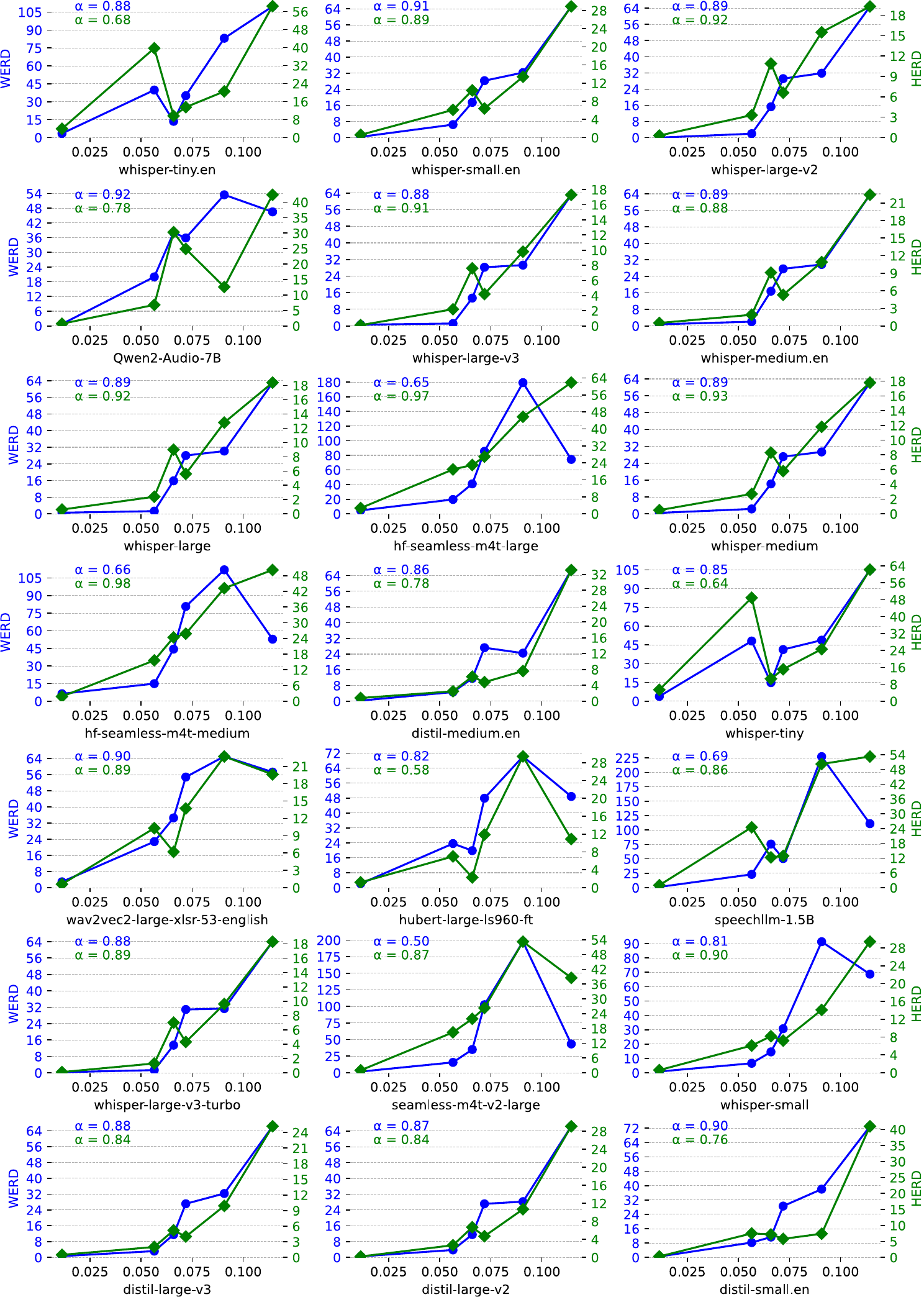}
\caption{ Degradation in WER (blue, left y-axis) and HER (green, right y-axis) w.r.t distribution shift (x-axis),
measured using Central Moment Discrepancy (CMD) for all models.}
    \label{fig:werd_herd_with_shift_all}
\end{figure*}

We provide a detailed analysis of the experimental results, focusing on the highest and lowest performing models across the study. To offer a comprehensive overview, we present Hallucination Error Rate (HER) and Word Error Rate (WER) across domain shifts for all models, as shown in~\ref{fig:werd_herd_with_shift_all}. 

Additionally, we include fine-grained error analysis in Table~\ref{tab:results_benchmark}, which highlights the differences between coarse-grained and fine-grained error categorization.

We also calculate the HER to WER ratio and present the numbers in Table~\ref{tab:results_ratio}. Robust models would exhibit a smaller gap between hallucination and non-hallucination errors. 

We provide results for transducer-based models along with Canary-1B in Table~\ref{tab:additional_models}.

Furthermore, we present the percentage of non-hallucination errors across datasets and models, categorizing them into Phonetic (P), Oscillation (O), and Language (L) errors in Table~\ref{tab:non_hallucination_error_analysis}. This analysis provides deeper insights into the types of errors that are most frequent and their distribution across different dataset-model combinations.

\begin{figure}
    \centering
    \includegraphics[width=1.0\linewidth]{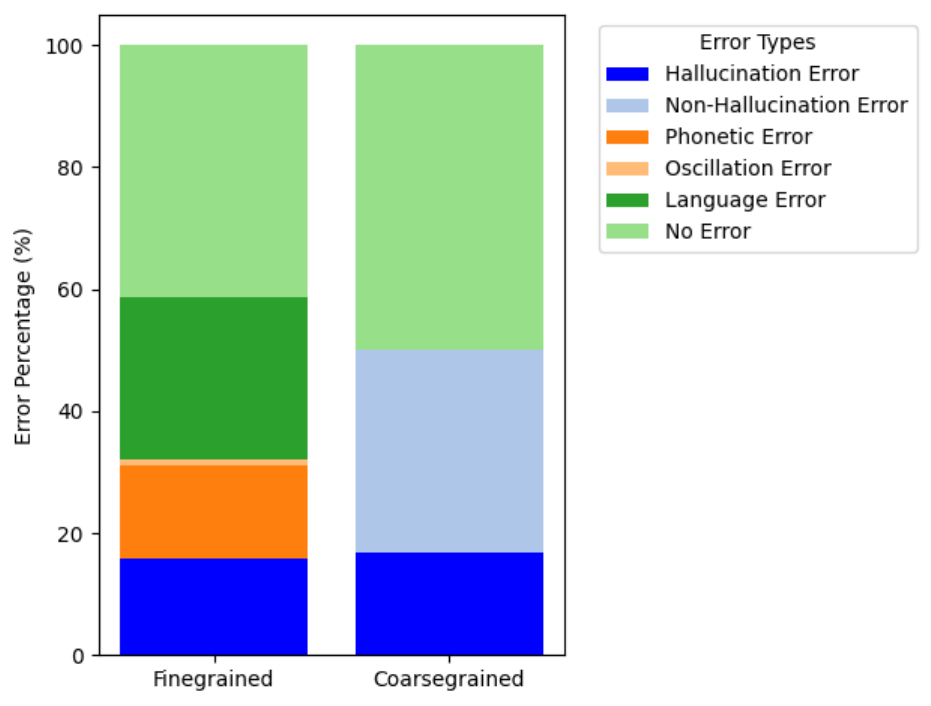}
    \caption{Finegrained vs coarsegrained error rate distribution averaged across all models and datasets.}
    \label{fig:pie_finegrained}
\end{figure}

We also highlight the overall distribution across all datasets and the robustness of both levels (coarse-grained and fine-grained) in correctly identifying hallucination in Figure~\ref{fig:pie_finegrained}. These results underscore the importance of considering both hallucination and non-hallucination errors when evaluating ASR systems, as well as the need for domain-specific adaptations to enhance robustness.

\subsubsection{LLM Evaluation}

We recognize that relying on GPT-4o for hallucination detection introduces the possibility that the evaluation model itself may produce incorrect labels, leading to false positives or negatives that could skew the assessment of ASR models. Given the length of our prompt (as shown in Figures~\ref{fig:coarsegrained_prompt} and~\ref{fig:finegrained_prompt}), this also raises concerns about the ``lost in the middle'' effect~\cite{liu2023lostmiddlelanguagemodels}, which can further increase the risk of hallucination during evaluation.

To mitigate these challenges, our evaluation framework restricts GPT-4o's output to only the final classification decision, explicitly avoiding chain-of-thought generation that could introduce errors~\cite{kim2024prometheusinducingfinegrainedevaluation}. This is achieved through carefully designed prompts and a deterministic decoding strategy (temperature=0.0), ensuring consistent and reproducible outputs. Consequently, GPT-4o-mini consistently produces only the classification decision, minimizing the likelihood of introducing evaluation-induced hallucinations.

Our findings show strong alignment between GPT-4o-mini's evaluations and human judgments, with minimal false positives or false negatives. While such errors are inevitable in any classifier short of a perfect oracle, they are substantially less frequent in our approach compared to heuristic-based methods, which often misclassify phonetic and oscillation errors as hallucinations. False negatives are particularly rare, and most false positives arise from phonetic or minor oscillation errors—very rarely (less than 1\%) from semantically correct transcriptions.

\paragraph{Effect of Text Normalization}

We also examine the impact of text normalization that we use to post-process the text for error rate calculation and hallucination detection. While normalization removes filler words and disfluencies—potentially important for applications requiring verbatim transcripts~\cite{Ma_2023, talafha2023nshotbenchmarkingwhisperdiverse}—we find that it has negligible impact on hallucination classification. We report the results in Table~\ref{tab:normalization_coarsegrained} and~\ref{tab:normalization_finegrained} for coarse-grained and fine-grained evaluations, respectively.

In our evaluation of 500 examples, we observe no transitions from ``No Error'' to ``Hallucination Error,'' and only about a 1\% transition from ``Hallucination Error'' to ``Non-Hallucination Error'' across both coarse-grained and fine-grained evaluations.

\begin{table}[h]
\renewcommand{\arraystretch}{1}
\centering
\resizebox{\linewidth}{!}{%
\begin{tabular}{ccc}
\toprule
\textbf{From (Normalized)} & \textbf{To (Orthographic)} & \textbf{Count} \\
\midrule
HE & NE  & 2 \\
HE & NHE & 7 \\
NE & NHE & 1 \\
NHE & HE & 2 \\
\bottomrule
\end{tabular}
}
\caption{Transitions from normalized to orthographic transcriptions (coarse-grained) in our 500-example evaluation. Abbreviations: HE - Hallucination Error, NE - No Error, NHE - Non-Hallucination Error.}
\label{tab:normalization_coarsegrained}
\end{table}

\begin{table}[h]
\renewcommand{\arraystretch}{1}
\centering
\resizebox{\linewidth}{!}{%
\begin{tabular}{ccc}
\toprule
\textbf{From (Normalized)} & \textbf{To (Orthographic)} & \textbf{Count} \\
\midrule
HE & NE          & 1 \\
HE & OE          & 1 \\
HE & PE          & 5 \\
HE & HE          & 1 \\
OE & HE          & 1 \\
PE & HE          & 2 \\
PE & LE          & 1 \\
\bottomrule
\end{tabular}
}
\caption{Transitions from normalized to orthographic transcriptions (fine-grained) in our 500-example evaluation. Abbreviations: HE - Hallucination Error, NE - No Error, OE - Oscillation Error, PE - Phonetic Error, LE - Language Error.}
\label{tab:normalization_finegrained}
\end{table}

The overall agreement scores of 0.9324 (coarse-grained) and 0.9495 (fine-grained) further underscore the robustness of our evaluation framework, confirming that normalization does not compromise the integrity of hallucination detection. These findings demonstrate that for our use case, normalization preserves semantic fidelity without introducing artifacts that would affect hallucination classification, ensuring that our evaluation remains consistent and reliable even when ``disfluencies`` are removed.

\paragraph{Cost Analysis}

We acknowledge that using proprietary models such as GPT-4o-mini incurs significant costs, which may limit scalability. To address this concern, we conduct a detailed cost analysis comparing our LLM-based evaluation framework to a human evaluation baseline, which serves as an upper bound for cost. This analysis scales to 1M segments, each approximately 20 seconds in duration, totaling roughly 5,000 hours.

For the LLM-based evaluation, the combined fine- and coarse-grained prompts take approximately 1,000 tokens per example (upper bound), with most input tokens cached. The error class outputs typically require fewer than five tokens on average. At the time of evaluation, the pricing is \$0.075 per million input tokens and \$0.600 per million output tokens. Scaling this setup to 1M segments:

\begin{itemize}
    \item Input cost: 1,000M tokens $\times$ \$0.075/1M = \$75.
    \item Output cost: 5M tokens $\times$ \$0.600/1M = \$3.
\end{itemize}
The total upper-bound cost is therefore \$78. With batch inference, we expect this could be reduced by 50\%, making it even more cost-effective.

In contrast, our human evaluation baseline requires approximately 10 minutes to evaluate 50 examples. Scaling to 1M examples:
\begin{itemize}
    \item Human evaluation time: (10 minutes / 50 examples) $\times$ 1M = 200,000 minutes ($\approx$ 3,333 hours).
    \item Human evaluation cost (at \$10/hour): $\approx$ \$30,000.
\end{itemize}

Overall, this analysis demonstrates that our LLM-based evaluation framework is approximately 375 times cheaper than human evaluation, underscoring its scalability and cost-effectiveness for large-scale evaluation tasks.

\subsubsection{Synthetic Shift}\label{appsubsubsec:synthetic_shift}
We provide results for the additional synthetic perturbations in Table~\ref{tab:additional_perturbations}.

\begin{table*}[ht!]
\centering
\resizebox{\textwidth}{!}{%
\begin{tabular}{l|cccc|cccc|cccc|cccc|cccc}
\toprule
\multicolumn{1}{l|}{\multirow{2}{*}{Model}} &
\multicolumn{4}{c|}{Cross-lingual Noise} &
\multicolumn{4}{c|}{Distortion} &
\multicolumn{4}{c|}{Echo} &
\multicolumn{4}{c|}{Pub Noise} &
\multicolumn{4}{c}{Reverberation} \\
& WER & CER & cHER & fHER & WER & CER & cHER & fHER & WER & CER & cHER & fHER & WER & CER & cHER & fHER & WER & CER & cHER & fHER \\
\midrule
distil-medium.en & 25.95 & 18.06 & 14.60 & 14.30 & 20.89 & 14.26 & 8.50 & 7.80 & 32.42 & 20.74 & 20.90 & 19.00 & 41.49 & 28.10 & 30.00 & 28.20 & 23.98 & 16.00 & 11.00 & 10.60 \\
whisper-large-v3 & 22.93 & 16.43 & 12.30 & 10.90 & 17.71 & 12.79 & 6.40 & 6.40 & 32.74 & 24.14 & 12.80 & 12.10 & 36.31 & 27.24 & 23.60 & 23.10 & 21.21 & 15.16 & 8.30 & 9.20 \\
wav2vec2-large-xlsr-53-english & 53.33 & 35.74 & 25.60 & 28.80 & 34.65 & 18.69 & 9.90 & 9.60 & 83.62 & 53.84 & 39.50 & 47.10 & 52.61 & 31.51 & 30.10 & 28.10 & 42.38 & 24.41 & 13.50 & 15.90 \\
\bottomrule
\end{tabular}%
}
\caption{WER, CER, cHER, and fHER of different models evaluated across various perturbation scenarios: Cross-lingual Noise, Distortion, Echo, Pub Noise, and Reverberation. Abbreviations: WER/CER - Word/Character Error Rate; cHER - Coarse-grained Hallucination Error Rate; fHER - Fine-grained Hallucination Error Rate.}

\label{tab:additional_perturbations}
\end{table*}



\begin{table*}[ht!]
\centering
\renewcommand{\arraystretch}{1}
\begin{tabular}{@{}p{2.5cm}cc@{}}
\toprule
\textbf{Error Type} & \textbf{Reference} & \textbf{Hypothesis} \\
\midrule
\textbf{Phonetic} & It's hard to recognize speech & It's hard to wreck a nice beach \\
\textbf{Hallucination} & There are more coming & Now move to me \\
\textbf{Language} & ripped ocean jumper &  Rips Ocean Jumper\\
\textbf{Oscillation} & oh so it's uh yeah &  Oh, I see, I see, I see, I see, I see, I see,\\
\bottomrule
\end{tabular}
\caption{Examples of different ASR error types, including phonetic confusions, hallucinations, language errors, and oscillation artifacts.}
\label{tab:hallucination_examples}
\end{table*}

\begin{table*}[h!]
\centering
\begin{tabular}{p{3cm}p{2.5cm}p{2.5cm}p{6cm}}
\toprule
\textbf{Name} & \textbf{Domain} & \textbf{Recording Conditions} & \textbf{Description} \\ \midrule
\textbf{LibriSpeech} & Speech Recognition & High-quality, read speech from audiobooks & A corpus of approximately 1,000 hours of 16kHz read English speech, derived from LibriVox audiobooks, segmented and aligned for ASR tasks. \\
\textbf{GLOBE} & Accented Speech & Close-talk microphone & Contains utterances from 23,519 speakers and covers 164 accents worldwide, recorded in close-talk microphone conditions. \\
\textbf{Supreme-court} & Legal & Diverse acoustic conditions (audiobooks, podcasts, YouTube) & A multi-domain, multi-style speech recognition corpus incorporating diverse acoustic and linguistic conditions, sourced from audiobooks, podcasts, and YouTube. \\
\textbf{SPGISpeech} & Finance & Corporate earnings calls (professional transcription) & Contains 5,000 hours of professionally transcribed audio from corporate earnings calls, featuring both spontaneous and narrated speaking styles. \\
\textbf{Adversarial} & Synthetic & Corporate earnings calls & Includes multiple splits with utterances modified using adversarial noise of varying radii (0.04 and 0.015) and combined with Room Impulse Response (RIR) noise. \\
\textbf{AMI (IHM)} & Meetings & Multi-device meeting environment & The AMI Meeting Corpus is a 100-hour dataset of English meeting recordings, featuring multimodal data synchronized across various devices. \\
\textbf{SLUE - VoxCeleb} & Conversational & YouTube video extracts (conversational) & Consists of single-sided conversational voice snippets extracted from YouTube videos, originally designed for speaker recognition. \\
\textbf{Primock57} & Medical & Mock consultations by clinicians & Contains mock consultations conducted by seven clinicians and 57 actors posing as patients, representing a diverse range of ethnicities, accents, and ages. \\
\textbf{BERSt} & Home Environment & Home recordings using smartphones & A collection of speech data recorded in home environments using various smartphone microphones, with participants from diverse regions. \\
\textbf{ATCOsim} & Aviation & Real-time air traffic control simulations & A specialized database containing ten hours of English speech from ten non-native speakers, recorded during real-time air traffic control simulations. \\ \bottomrule
\end{tabular}
\caption{Speech Datasets for ASR, categorized by domain, recording conditions, and description.}
\label{tab:datasets}
\end{table*}


\begin{table*}[h!]
\renewcommand{\arraystretch}{1.2}
\centering
\resizebox{\textwidth}{!}{%
\begin{tabular}{p{0.2\textwidth} p{0.15\textwidth} p{0.2\textwidth} p{0.45\textwidth}}
\hline
\textbf{Model Type and Models} & \textbf{Parameters} & \textbf{Architecture} & \textbf{Pre-Training Objective and Training Data} \\ \hline

\textbf{wav2vec2}~\cite{baevski2020wav2vec} \newline 
-- wav2vec2-large-xlsr-53-english & 
315M & 
7-Conv (Kernel 10/3/3/3/3/2/2) + 24-Trans & 
Self-supervised pre-training on raw audio via contrastive loss. \newline 
Training data: Common Voice 6.1 (53 languages). \\ \hline

\textbf{hubert}~\cite{hsu2021hubert}\newline  
-- hubert-large-ls960-ft & 
316M & 
7-Conv (Kernel 10/3/3/3/3/2/2) + 24-Trans & 
Masked prediction pre-training. \newline 
Training data: Libri-Light (60k hours). \\ \hline

\textbf{seamless}~\cite{seamless2023} \newline 
-- hf-seamless-m4t-large \newline 
-- hf-seamless-m4t-medium & 
2.3B \newline 
1.2B & 
UnitY2 (Enc-Dec + Text Decoder) & 
Multilingual ASR/translation. \newline 
Training data: 443k hours of aligned speech-text (29 languages). \\ \hline

\textbf{speechllm}~\cite{Rajaa_SpeechLLM_Multi-Modal_LLM} \newline 
-- speechllm-1.5B & 
1.5B & 
HubertX encoder + TinyLlama decoder & 
Audio-text alignment via multi-task learning. \newline 
Training data: Proprietary ASR datasets. \\ \hline

\textbf{whisper}~\cite{radford2022robustspeechrecognitionlargescale} \newline 
-- whisper-large-v3 \newline 
-- distil-large-v3 \newline 
-- whisper-large-v2 \newline 
-- whisper-large-v3-turbo \newline 
-- distil-large-v2 \newline 
-- whisper-large \newline 
-- whisper-tiny \newline 
-- whisper-tiny.en \newline 
-- whisper-medium \newline 
-- whisper-medium.en \newline 
-- distil-medium.en \newline 
-- distil-small.en \newline 
-- whisper-small \newline 
-- whisper-small.en & 
1.55B \newline 
756M \newline 
769M \newline 
244M \newline
39M &
2-Conv (Kernel 3x3, stride 2) + 32-Trans (large) \newline 
2-Conv + 24-Trans (medium) \newline 
2-Conv + 12-Trans (small) & 
Multilingual ASR/translation. \newline 
Pre-training: 680k hours of web-crawled audio. \\ \hline

\textbf{Qwen}~\cite{chu2024qwen2}\newline 
-- Qwen2-Audio-7B & 
7B & 
Audio encoder + QwenLM decoder & 
Multi-task pretraining (ASR, TTS, alignment). \newline 
Training data: 3M audio-text pairs. \\ \hline

\end{tabular}%
}
\caption{Model architectures, parameters, and training details. Whisper variants include convolutional layers for spectrogram downsampling.}
\label{tab:models}
\end{table*}

\begin{figure*}[h]
    \centering
    \tcbset{colframe=black, colback=gray!10, arc=5mm}
    \begin{tcolorbox}
    \small
    \textbf{You are a classifier trained to detect and categorize specific transcription errors produced by a speech recognition system.} The possible categories are:

    1. \textbf{Hallucination Error}: The output contains fabricated, contradictory, or invented information that is not supported by the ground truth. This includes:
       - \textbf{Fabricated Content}: Words or phrases entirely absent in the ground truth.
       - \textbf{Meaningful Contradictions}: Significant changes in the meaning from the ground truth.
       - \textbf{Invented Context}: Introduction of details or context not present in the ground truth.  
       - \textbf{Note}: These errors involve fabrication of new information or significant distortion of meaning, beyond grammatical or structural mistakes.

    2. \textbf{Non-Hallucination Error}: Errors that do not involve fabrication or significant contradictions of the ground truth. These include:
       - \textbf{Phonetic Errors}: Substitutions of phonetically similar words or minor pronunciation differences.
       - \textbf{Structural or Language Errors}: Grammatical, syntactic, or structural issues that make the text incoherent or incorrect (e.g., incorrect verb tenses, subject-verb agreement problems, omissions, or insertions).
       - \textbf{Oscillation Errors}: Repetitive, nonsensical patterns or sounds that do not convey linguistic meaning (e.g., "ay ay ay ay").
       - \textbf{Other Non-Hallucination Errors}: Errors that do not fit the above subcategories but are not hallucinations.

    3. \textbf{No Error}: The generated output conveys the same meaning as the ground truth, even if the phrasing, grammar, or structure differs. Minor differences in wording, phrasing, or grammar that do not alter the intended meaning are acceptable. \\

    \textbf{Input Format:} \\
    \textbf{Ground Truth}: The original, accurate text provided. \\
    \textbf{Generated Output}: The text produced by the speech recognition system. \\

    \textbf{Output Format:}
    Classify the input text pairs into one of the following: \\
    Non-Hallucination Error \\
    Hallucination Error \\
    No Error \\

    \textbf{Examples:} \\
    \textbf{Example 1:} \\
    Ground Truth: "A millimeter roughly equals one twenty-fifth of an inch." \\
    Generated Output: "Miller made her roughly one twenty-fifths of an inch." \\
    \textbf{Output}: Non-Hallucination Error \\

    \textbf{Example 2:} \\
    Ground Truth: "Indeed, ah!" \\
    Generated Output: "Ay ay indeed ay ay ay ay ay ay." \\
    \textbf{Output}: Non-Hallucination Error \\

    \textbf{Example 3:} \\
    Ground Truth: "Captain Lake did not look at all like a London dandy now." \\
    Generated Output: "Will you let Annabel ask her if she sees what it is you hold in your arms again?" \\
    \textbf{Output}: Hallucination Error \\

    \textbf{Example 4:} \\
    Ground Truth: "The patient was advised to take paracetamol for fever and rest for two days." \\
    Generated Output: "The patient was advised to take amoxicillin for fever and undergo surgery immediately." \\
    \textbf{Output}: Hallucination Error \\

    \textbf{Example 5:} \\
    Ground Truth: "I need to book a flight to New York." \\
    Generated Output: "I need to book ticket to New York." \\
    \textbf{Output}: No Error \\

    \textbf{Example 6:} \\
    Ground Truth: "She went to the store yesterday." \\
    Generated Output: "She went to the shop yesterday." \\
    \textbf{Output}: No Error \\

    \textbf{Instruction:}  
    You must produce only the classification as the output. Do not include explanations, reasoning, or additional information. \\

    \textbf{Input:}  
    Ground Truth: "\{ground\_truth\}"  
    Generated Output: "\{output\}"  

    \textbf{Output:} \{\{insert your classification here\}\}
    \end{tcolorbox}
    \caption{Coarsegrained error detection prompt. The task is to classify transcription errors produced by an ASR model into one of three categories: \textit{Non-Hallucination Error}, \textit{Hallucination Error}, or \textit{No Error}.}
    \label{fig:coarsegrained_prompt}
\end{figure*}
\begin{figure*}[h]
    \centering
    \tcbset{colframe=black, colback=gray!10, arc=5mm}
    \begin{tcolorbox}
    \small
    \textbf{You are a classifier trained to detect and categorize specific transcription errors produced by a speech recognition system.} The possible categories are:

    1. \textbf{Phonetic Error}: The output contains substitutions of phonetically similar words that do not match the ground truth and do not introduce broader grammatical or structural issues. These errors typically involve misrecognition of similar-sounding words or minor pronunciation differences.

    2. \textbf{Oscillation Error}: The output includes repetitive, nonsensical patterns or sounds that do not convey linguistic meaning (e.g., "ay ay ay ay").

    3. \textbf{Hallucination Error}: The output contains fabricated, contradictory, or invented information that is not supported by the ground truth. This includes:
       - \textbf{Fabricated Content}: Words or phrases entirely absent in the ground truth.
       - \textbf{Meaningful Contradictions}: Significant changes in the meaning from the ground truth.
       - \textbf{Invented Context}: Introduction of details or context not present in the ground truth.  
       - \textbf{Note}: These errors involve fabrication of new information or significant distortion of meaning, beyond grammatical or structural mistakes.

    4. \textbf{Language Error}: The output includes grammatical, syntactic, or structural issues that make the text incoherent or linguistically incorrect. This category encompasses errors such as:
       - Incorrect verb tenses or subject-verb agreement problems.
       - Sentence fragments or incomplete structures.
       - Omissions or insertions of words that do not fabricate new context.
       - Incomplete sentences or phrases that do not convey the intended meaning as ground truth.  
       - \textbf{Note}: Incomplete sentences or phrases are classified as Language Errors only when they do not fabricate new meaning or deviate from the intent of the ground truth.

    5. \textbf{No Error}: The generated output conveys the same meaning as the ground truth, even if the phrasing, grammar, or structure differs. Minor differences in wording, phrasing, punctuation, or casing that do not alter the intended meaning are not considered errors.  
       - \textbf{Note}: Minor omissions, such as missing articles, are acceptable as long as they do not change the meaning of the ground truth. \\

    \textbf{Input Format:} \\
    \textbf{Ground Truth}: The original, accurate text provided. \\
    \textbf{Generated Output}: The text produced by the speech recognition system. \\

    \textbf{Output Format:}
    Classify the input text pairs into one of the following: \\
    Phonetic Error \\
    Oscillation Error \\
    Hallucination Error \\
    Language Error \\
    No Error \\

    \textbf{Examples:} \\
    \textbf{Example 1:} \\
    Ground Truth: "A millimeter roughly equals one twenty-fifth of an inch." \\
    Generated Output: "Miller made her roughly one twenty-fifths of an inch." \\
    \textbf{Output:} Phonetic Error \\
    
    \textbf{Example 2:} \\
    Ground Truth: "I will go to New York City!" \\
    Generated Output: "Ay ay ay ay ay ay ay ay." \\
    \textbf{Output:} Oscillation Error \\
    
    \textbf{Example 3:} \\
    Ground Truth: "Captain Lake did not look at all like a London dandy now." \\
    Generated Output: "Will you let Annabel ask her if she sees what it is you hold in your arms again?" \\
    \textbf{Output:} Hallucination Error \\
    
    \textbf{Example 4:} \\
    Ground Truth: "The cat is chasing the mouse." \\
    Generated Output: "The cat chased by the mouse." \\
    \textbf{Output:} Language Error \\
    
    \textbf{Example 5:} \\
    Ground Truth: "I need to book a flight to New York." \\
    Generated Output: "I need to book ticket to New York." \\
    \textbf{Output:} No Error \\

    \textbf{Your Task:}
    Classify the input into one of the five categories.

    \textbf{Instruction:}  
    You must produce only the classification as the output. Do not include explanations, reasoning, or additional information. \\

    \textbf{Input:}  
    Ground Truth: "\{ground\_truth\}"  
    Generated Output: "\{output\}"  

    \textbf{Output:} \{\{insert your classification here\}\}
    \end{tcolorbox}
    \caption{Finegrained error detection prompt. The task is to classify transcription errors produced by an ASR model into one of five categories: \textit{Phonetic Error}, \textit{Oscillation Error}, \textit{Hallucination Error}, \textit{Language Error}, or \textit{No Error}.}
    \label{fig:finegrained_prompt}
\end{figure*}


\begin{table*}[ht!]
\centering
\resizebox{\textwidth}{!}{%
\begin{tabular}{l|cccc|cccc|cccc|cccc|cccc}
\toprule
\multicolumn{1}{l|}{\multirow{2}{*}{Model}} &
\multicolumn{4}{c|}{BERSt} &
\multicolumn{4}{c|}{Primock57} &
\multicolumn{4}{c|}{Adversarial} &
\multicolumn{4}{c|}{Salt-multispeaker-eng} &
\multicolumn{4}{c}{Supreme-court-speech} \\
& WER & CER & cHER & fHER & WER & CER & cHER & fHER & WER & CER & cHER & fHER & WER & CER & cHER & fHER & WER & CER & cHER & fHER \\
\midrule
canary-1b & 37.4 & 37.4 & 18.8 & 14.7 & 18.0 & 17.2 & 6.0 & 6.6 & 27.9 & 27.5 & 46.3 & 41.6 & 3.3 & 3.3 & 2.1 & 1.6 & 16.0 & 15.7 & 18.9 & 21.3 \\
parakeet-rnnt-1.1b & 26.1 & 27.7 & 16.4 & 13.9 & 17.0 & 16.2 & 5.7 & 6.3 & 17.5 & 17.2 & 29.8 & 30.6 & 3.1 & 3.1 & 1.6 & 2.1 & 15.5 & 15.3 & 17.4 & 18.8 \\
parakeet-tdt-1.1b & 27.7 & 28.6 & 12.6 & 11.1 & 15.8 & 15.0 & 5.5 & 6.4 & 19.4 & 19.1 & 32.2 & 30.2 & 3.3 & 3.3 & 1.1 & 1.6 & 15.7 & 15.5 & 15.7 & 19.1 \\
\bottomrule
\end{tabular}%
}
\caption{WER, CER, cHER, and fHER for different models across datasets.}
\label{tab:additional_models}
\end{table*}

\begin{table*}[ht!]
\resizebox{\textwidth}{!}{%
\begin{tabular}{lccccccccccccc}
\toprule
\multicolumn{1}{l}{\multirow{1}{*}{Model}} & 
\multicolumn{1}{c}{SPGI} & 
\multicolumn{1}{c}{BERSt} & 
\multicolumn{1}{c}{ATCOsim} & 
\multicolumn{1}{c}{ADV} & 
\multicolumn{1}{c}{AMI} & 
\multicolumn{1}{c}{SLU} & 
\multicolumn{1}{c}{SNIPS} & 
\multicolumn{1}{c}{SC} & 
\multicolumn{1}{c}{GLOBE} & 
\multicolumn{1}{c}{SALT} & 
\multicolumn{1}{c}{LS\_Noise} & 
\multicolumn{1}{c}{LS} & 
\multicolumn{1}{c}{Primock57} \\
\midrule
whisper-large-v3 & 3.4/0.9 & 32.4/13.2 & 65.3/13.1 & 33.3/47.1 & 23.4/11.3 & 15.5/13.4 & 8.2/0.5 & 18.8/14.8 & 3.4/1.9 & 3.0/1.0 & 2.6/0.4 & 2.2/0.5 & 19.2/4.8 \\ 
wav2vec2-large-xlsr-53-english & 19.6/2.9 & 64.0/13.9 & 63.0/11.9 & 100.1/95.7 & 53.0/22.8 & 43.4/14.6 & 12.4/0.9 & 32.6/17.9 & 27.0/9.3 & 17.0/2.1 & 9.0/0.8 & 6.5/0.1 & 47.9/15.8 \\ 
hf-seamless-m4t-large & 14.7/4.5 & 58.9/29.5 & 76.6/55.1 & 61.2/71.4 & 63.7/43.2 & 44.0/27.5 & 7.4/2.6 & 34.2/30.8 & 19.2/19.2 & 4.2/1.0 & 6.5/2.9 & 3.4/0.3 & 44.5/25.8 \\ 
speechllm-1.5B & 11.5/4.1 & 68.5/31.8 & 121.4/38.3 & 95.3/92.5 & 127.3/52.3 & 83.8/19.3 & 10.8/2.8 & 41.3/36.5 & 27.9/21.8 & 9.5/4.2 & 10.9/4.3 & 11.4/4.2 & 41.7/17.3 \\ 
whisper-medium & 3.7/1.1 & 34.5/14.5 & 65.6/14.4 & 42.9/58.4 & 23.2/11.4 & 17.4/13.6 & 8.6/1.1 & 18.7/14.0 & 5.3/2.9 & 5.0/3.7 & 3.3/0.8 & 3.1/0.2 & 20.6/5.9 \\ 
distil-large-v2 & 4.1/0.9 & 38.0/14.7 & 69.5/22.4 & 45.6/60.4 & 22.1/13.5 & 16.0/13.4 & 9.2/0.8 & 18.9/15.0 & 6.7/3.0 & 5.2/1.0 & 3.6/0.5 & 3.4/0.3 & 19.2/5.9 \\ 
hubert-large-ls960-ft & 12.4/2.0 & 58.5/11.3 & 50.0/6.4 & 109.8/100.0 & 44.4/28.1 & 21.3/6.5 & 12.6/1.3 & 30.2/23.8 & 23.4/6.5 & 18.7/4.2 & 3.6/2.1 & 2.2/0.1 & 32.2/11.7 \\ 
distil-medium.en & 4.6/0.6 & 39.3/14.8 & 71.3/26.8 & 45.8/58.8 & 23.6/10.6 & 15.6/11.3 & 9.7/1.1 & 20.0/15.2 & 8.5/1.9 & 7.4/3.1 & 4.3/0.9 & 4.2/0.5 & 21.0/5.4 \\ 
distil-small.en & 4.6/1.0 & 46.8/17.9 & 77.0/32.7 & 54.3/68.6 & 24.2/10.9 & 15.4/13.1 & 11.3/1.7 & 21.5/18.5 & 11.7/6.4 & 9.0/4.7 & 4.1/0.8 & 4.0/0.4 & 21.4/6.8 \\ 
whisper-medium.en & 4.3/1.5 & 34.2/15.2 & 66.6/16.2 & 43.3/58.0 & 23.0/11.2 & 19.4/16.3 & 8.4/1.4 & 21.3/16.3 & 4.8/1.7 & 5.7/4.2 & 3.5/0.7 & 3.1/0.4 & 20.6/6.0 \\ 
whisper-small.en & 4.1/1.3 & 38.7/17.5 & 68.8/19.3 & 50.9/69.8 & 24.5/13.5 & 20.8/15.9 & 9.4/1.1 & 20.9/16.5 & 9.6/4.8 & 7.2/4.7 & 3.7/0.9 & 3.6/0.4 & 21.5/6.7 \\ 
speecht5\_asr & 25.8/28.1 & 108.1/32.3 & 81.7/57.4 & 117.2/100.0 & 462.5/25.1 & 129.4/21.9 & 24.6/7.8 & 156.7/43.9 & 60.1/54.1 & 53.9/28.3 & 13.9/14.1 & 6.0/0.8 & 53.9/41.2 \\ 
hf-seamless-m4t-medium & 13.2/5.1 & 57.9/29.1 & 52.7/40.9 & 51.4/63.5 & 57.0/41.9 & 50.3/25.3 & 8.8/1.6 & 36.0/32.6 & 15.9/14.2 & 6.4/2.1 & 8.9/3.0 & 3.7/0.4 & 46.1/24.5 \\ 
whisper-tiny & 8.8/4.3 & 122.1/37.2 & 110.3/60.3 & 88.0/85.9 & 40.3/25.3 & 22.5/15.4 & 15.6/4.7 & 38.6/28.3 & 54.7/47.6 & 20.0/13.1 & 10.8/6.7 & 7.6/1.6 & 32.8/15.8 \\ 
whisper-large & 3.7/1.1 & 48.1/12.8 & 65.7/14.2 & 37.2/51.4 & 22.6/12.8 & 18.1/15.5 & 8.5/0.9 & 18.6/14.5 & 4.2/1.8 & 4.0/2.6 & 3.1/0.5 & 3.0/0.2 & 20.0/6.0 \\ 
whisper-large-v2 & 4.3/0.9 & 34.1/14.5 & 67.1/15.1 & 38.9/54.1 & 24.1/14.3 & 18.2/15.9 & 8.4/0.5 & 23.6/15.6 & 4.4/3.0 & 3.2/1.6 & 2.7/0.6 & 3.0/0.2 & 20.0/6.4 \\ 
whisper-large-v3-turbo & 3.4/0.9 & 31.7/11.7 & 66.2/13.5 & 34.5/47.1 & 23.8/10.6 & 15.7/14.4 & 7.8/0.6 & 18.5/14.4 & 3.9/1.1 & 4.7/1.6 & 2.7/0.3 & 2.5/0.5 & 20.8/4.8 \\ 
whisper-tiny.en & 6.9/3.0 & 75.4/32.5 & 112.3/57.9 & 80.1/82.7 & 38.4/20.8 & 19.4/15.1 & 14.0/4.1 & 38.0/28.0 & 42.1/37.9 & 19.4/12.6 & 9.4/5.1 & 6.1/0.8 & 31.3/14.8 \\ 
distil-large-v3 & 3.7/0.7 & 33.7/12.0 & 69.2/18.0 & 38.6/51.0 & 23.4/11.7 & 14.1/12.9 & 8.8/0.9 & 19.5/16.7 & 5.6/1.3 & 5.0/2.1 & 3.3/0.6 & 2.8/0.3 & 19.1/5.6 \\ 
whisper-small & 4.3/1.2 & 42.1/17.7 & 73.4/23.6 & 77.0/72.5 & 39.8/14.7 & 18.2/14.2 & 9.6/1.4 & 23.4/17.6 & 10.0/4.4 & 7.1/4.2 & 4.3/0.4 & 3.7/0.3 & 22.1/7.4 \\ 
Qwen2-Audio-7B & 4.6/2.7 & 36.3/15.6 & 44.8/35.7 & 31.7/46.7 & 35.7/14.9 & 47.4/32.1 & 5.5/1.3 & 35.3/37.1 & 23.3/7.0 & 5.9/5.8 & 2.3/1.3 & 2.0/0.7 & 25.5/22.8 \\ 
seamless-m4t-v2-large & 15.9/5.4 & 55.2/25.8 & 43.5/31.0 & 50.5/67.5 & 75.1/50.6 & 45.9/21.3 & 6.0/1.6 & 34.7/24.8 & 14.9/14.9 & 5.3/1.6 & 3.6/1.5 & 2.7/0.4 & 37.6/23.5 \\ 
\bottomrule
\end{tabular}
}
\caption{Character Error Rate (CER) and hallucination error rate (HER) across models and datasets. Values are presented as CER/HER.}
\label{tab:results_benchmark}
\end{table*}

\begin{table*}[ht!]
\resizebox{\textwidth}{!}{%
\begin{tabular}{lccccccccccccc}
\toprule
\multicolumn{1}{l}{\multirow{1}{*}{Model}} & 
\multicolumn{1}{c}{SPGI} & 
\multicolumn{1}{c}{BERSt} & 
\multicolumn{1}{c}{ATCOsim} & 
\multicolumn{1}{c}{ADV} & 
\multicolumn{1}{c}{AMI} & 
\multicolumn{1}{c}{SLU} & 
\multicolumn{1}{c}{SNIPS} & 
\multicolumn{1}{c}{SC} & 
\multicolumn{1}{c}{GLOBE} & 
\multicolumn{1}{c}{SALT} & 
\multicolumn{1}{c}{LS\_Noise} & 
\multicolumn{1}{c}{LS} & 
\multicolumn{1}{c}{Primock57} \\
\midrule
whisper-large-v3 & 0.12 & 0.49 & 0.27 & 1.49 & 0.43 & 0.51 & 0.11 & 0.82 & 0.74 & 0.34 & 0.16 & 0.14 & 0.23 \\ 
wav2vec2-large-xlsr-53-english & 0.05 & 0.29 & 0.31 & 0.96 & 0.43 & 0.14 & 0.10 & 0.45 & 0.38 & 0.12 & 0.07 & 0.00 & 0.29 \\
hf-seamless-m4t-large & 0.24 & 0.59 & 0.81 & 1.26 & 0.73 & 0.53 & 0.43 & 0.88 & 1.11 & 0.37 & 0.49 & 0.15 & 0.62 \\ 
speechllm-1.5B & 0.42 & 0.56 & 0.47 & 0.99 & 0.43 & 0.20 & 0.27 & 0.93 & 1.03 & 0.44 & 0.48 & 0.37 & 0.41 \\ 
whisper-medium & 0.05 & 0.47 & 0.28 & 1.47 & 0.53 & 0.50 & 0.16 & 0.77 & 0.58 & 0.74 & 0.28 & 0.13 & 0.30 \\ 
distil-large-v2 & 0.25 & 0.42 & 0.43 & 1.41 & 0.51 & 0.45 & 0.14 & 0.76 & 0.48 & 0.20 & 0.20 & 0.15 & 0.27 \\ 
hubert-large-ls960-ft & 0.11 & 0.24 & 0.22 & 0.91 & 0.66 & 0.11 & 0.10 & 0.68 & 0.30 & 0.20 & 0.37 & 0.05 & 0.37 \\ 
distil-medium.en & 0.13 & 0.44 & 0.48 & 1.40 & 0.36 & 0.45 & 0.14 & 0.73 & 0.40 & 0.49 & 0.40 & 0.21 & 0.27 \\ 
distil-small.en & 0.13 & 0.41 & 0.54 & 1.35 & 0.33 & 0.51 & 0.20 & 0.85 & 0.69 & 0.53 & 0.22 & 0.15 & 0.30 \\ 
whisper-medium.en & 0.23 & 0.55 & 0.34 & 1.40 & 0.49 & 0.49 & 0.18 & 0.72 & 0.47 & 0.64 & 0.26 & 0.13 & 0.28 \\ 
whisper-small.en & 0.22 & 0.51 & 0.42 & 1.43 & 0.56 & 0.51 & 0.13 & 0.78 & 0.67 & 0.80 & 0.24 & 0.08 & 0.31 \\ 
hf-seamless-m4t-medium & 0.33 & 0.56 & 0.96 & 1.31 & 0.77 & 0.50 & 0.24 & 0.87 & 1.01 & 0.33 & 0.26 & 0.13 & 0.57 \\ 
whisper-tiny & 0.41 & 0.34 & 0.58 & 1.02 & 0.65 & 0.55 & 0.35 & 0.82 & 0.93 & 0.68 & 0.66 & 0.22 & 0.51 \\ 
whisper-large & 0.16 & 0.33 & 0.28 & 1.50 & 0.57 & 0.50 & 0.12 & 0.80 & 0.59 & 0.65 & 0.23 & 0.03 & 0.28 \\ 
whisper-large-v2 & 0.19 & 0.52 & 0.29 & 1.48 & 0.65 & 0.62 & 0.13 & 0.66 & 0.81 & 0.33 & 0.22 & 0.10 & 0.34 \\ 
whisper-large-v3-turbo & 0.12 & 0.46 & 0.28 & 1.44 & 0.42 & 0.47 & 0.12 & 0.75 & 0.44 & 0.55 & 0.19 & 0.16 & 0.23 \\ 
whisper-tiny.en & 0.35 & 0.45 & 0.53 & 1.08 & 0.56 & 0.55 & 0.35 & 0.75 & 0.97 & 0.62 & 0.53 & 0.18 & 0.47 \\ 
distil-large-v3 & 0.03 & 0.41 & 0.37 & 1.45 & 0.44 & 0.40 & 0.11 & 0.78 & 0.43 & 0.32 & 0.27 & 0.14 & 0.23 \\ 
whisper-small & 0.14 & 0.46 & 0.40 & 0.98 & 0.36 & 0.47 & 0.14 & 0.74 & 0.64 & 0.67 & 0.21 & 0.08 & 0.34 \\ 
Qwen2-Audio-7B & 0.74 & 0.52 & 0.96 & 1.55 & 0.38 & 0.66 & 0.26 & 1.18 & 0.33 & 1.16 & 0.67 & 0.39 & 1.01 \\ 
seamless-m4t-v2-large & 0.41 & 0.55 & 0.90 & 1.42 & 0.71 & 0.49 & 0.29 & 0.71 & 1.13 & 0.40 & 0.41 & 0.19 & 0.71 \\ 
\bottomrule
\end{tabular}
}
\caption{Comparison of HER/WER ratio across models for all datasets.}
\label{tab:results_ratio}
\end{table*}


\begin{table*}[]
\resizebox{\textwidth}{!}{%
\begin{tabular}{cccccccccccccccccccccccccccccccccc}
\toprule
\multirow{2}{*}{Model} & \multicolumn{3}{c}{BERSt} & \multicolumn{3}{c}{GLOBE} & \multicolumn{3}{c}{LibriSpeech} & \multicolumn{3}{c}{Primock57} & \multicolumn{3}{c}{Adversarial} & \multicolumn{3}{c}{AMI} & \multicolumn{3}{c}{ATCOsim} & \multicolumn{3}{c}{SALT} & \multicolumn{3}{c}{SLUE} & \multicolumn{3}{c}{SPGI} & \multicolumn{3}{c}{SC} \\ 
                       & P       & O      & L      & P       & O      & L      & P         & O        & L        & P        & O       & L        & P         & O        & L        & P      & O     & L      & P        & O       & L      & P       & O      & L     & P       & O     & L      & P       & O      & L     & P     & O      & L     \\ \midrule
Q2A-7B         & 37.41   & 0.75   & 5.64   & 21.3    & 3.5    & 8      & 8.6       & 0.2      & 1.3      & 10.5     & 5.5     & 12.5     & 19.22     & 0.39     & 4.31     & 10     & 3.4   & 12.4   & 49.5     & 0.7     & 3.4    & 11      & 1.7    & 2.3   & 10.1    & 0.2   & 1.7    & 8.38    & 2.09   & 4.71  & 10.7  & 15.6   & 24.1  \\
dw-l-v2                & 35.53   & 0.19   & 5.08   & 22.7    & 0      & 4.7    & 17.7      & 0.1      & 5.4      & 10.2     & 0.7     & 14.8     & 16.08     & 0        & 13.73    & 7.7    & 0     & 13.7   & 57.9     & 0       & 2.8    & 12.5    & 5.7    & 4.5   & 15.9    & 0     & 4.2    & 15.18   & 0      & 3.66  & 11.8  & 18.5   & 27.2  \\
dw-l-v3                & 33.27   & 0      & 3.57   & 21.2    & 0      & 3.9    & 14.3      & 0.1      & 3.5      & 9.3      & 0.6     & 13.6     & 24.71     & 0        & 10.59    & 6.6    & 0     & 12.3   & 61.2     & 0       & 2.6    & 14.4    & 3.4    & 4.1   & 13.5    & 0     & 3.4    & 12.57   & 0      & 3.66  & 11.3  & 18.5   & 26.8  \\
dw-m.en                & 45.3    & 1.32   & 2.82   & 24.6    & 0      & 6.8    & 18        & 0        & 7.9      & 10.1     & 0.5     & 18.3     & 16.86     & 0.39     & 15.29    & 7.2    & 0     & 14.4   & 58.8     & 0.5     & 3.2    & 12.6    & 6.2    & 5.8   & 16.7    & 0     & 5.7    & 21.99   & 0      & 6.28  & 13    & 21.2   & 31.6  \\
dw-s.en                & 40.6    & 0.75   & 5.08   & 29.6    & 0.1    & 7.7    & 21.2      & 0.1      & 6.4      & 13.4     & 1.1     & 17.2     & 13.73     & 0        & 12.16    & 7.8    & 0     & 12.1   & 53.9     & 0.6     & 2.8    & 7.4     & 15.7   & 2.2   & 18      & 0.3   & 5.2    & 24.61   & 0      & 4.19  & 13    & 20.5   & 29.8  \\
sm4t-l                 & 41.73   & 0      & 4.32   & 17.7    & 0.3    & 2.4    & 20.3      & 0        & 5.1      & 7.9      & 1.7     & 13.9     & 7.06      & 1.96     & 9.02     & 6.4    & 0     & 14.8   & 29.6     & 1.6     & 4.2    & 1.2     & 21.8   & 0.2   & 20.2    & 0     & 2.3    & 7.33    & 0      & 2.09  & 4.6   & 15.3   & 25.3  \\
sm4t-m                 & 39.29   & 0      & 3.76   & 21.9    & 0.4    & 3.7    & 24.5      & 0.1      & 6.9      & 8.5      & 1.9     & 15.6     & 15.69     & 1.18     & 3.53     & 5.1    & 0     & 15.2   & 47.2     & 1       & 3.2    & 3.7     & 17.2   & 2.5   & 23.8    & 0     & 3.6    & 9.95    & 0      & 3.66  & 5.8   & 16.4   & 29    \\
hubert                 & 66.17   & 3.2    & 0.56   & 58.4    & 0.2    & 4.4    & 21.8      & 0.1      & 3.1      & 62.7     & 1.3     & 10.4     & 0         & 0        & 0        & 47.2   & 0     & 8.6    & 91.4     & 0.2     & 0.5    & 5.3     & 52.4   & 0.1   & 17.2    & 0     & 1      & 56.54   & 0      & 5.76  & 52.3  & 7.3    & 30.5  \\
sm4t-v2-l              & 42.29   & 0      & 3.38   & 15      & 0.2    & 3      & 17.9      & 0.2      & 2.6      & 9.1      & 2       & 16.2     & 13.33     & 0.39     & 2.75     & 4.1    & 0     & 17.6   & 49.1     & 1.6     & 4.9    & 2.8     & 20.6   & 0.9   & 17.4    & 0     & 2.2    & 6.28    & 0      & 2.62  & 8.9   & 13.8   & 26.4  \\
spllm-1.5B             & 50.56   & 1.13   & 1.32   & 39.9    & 0.9    & 6.7    & 42.9      & 0.3      & 4.5      & 24.4     & 6.2     & 20.7     & 3.14      & 1.57     & 1.96     & 6.67   & 0     & 17.5   & 55.51    & 3.81    & 1      & 1.3     & 2      & 0.3   & 38.8    & 0.5   & 3.5    & 29.32   & 0      & 2.62  & 22.3  & 13.82  & 33.6  \\
w2v2-large             & 72.74   & 0.19   & 0.75   & 64.6    & 0.2    & 3.5    & 46.6      & 0.1      & 9.7      & 55.3     & 1.2     & 14.1     & 1.96      & 2.35     & 0        & 50.2   & 0     & 7.8    & 87.2     & 0       & 0.2    & 6.7     & 36     & 0     & 35.5    & 0     & 8.8    & 58.12   & 0      & 3.66  & 43.7  & 9.4    & 37.7  \\
w-large                & 33.83   & 0.56   & 1.69   & 14.9    & 0      & 2.2    & 12.1      & 0        & 2.7      & 6.2      & 0.3     & 10.6     & 20        & 0.39     & 7.45     & 6.4    & 0     & 9.9    & 48.6     & 0       & 1.3    & 12.5    & 5.8    & 2.5   & 11.9    & 0     & 2.4    & 7.85    & 0      & 1.05  & 8.5   & 14.1   & 23.7  \\
w-l-v2                 & 36.84   & 0      & 2.63   & 14.3    & 0      & 2.4    & 11.3      & 0        & 2.1      & 5.1      & 1.1     & 10.2     & 16.86     & 0        & 9.8      & 6.2    & 0     & 11.2   & 45.5     & 0.2     & 1      & 10.3    & 3.4    & 1.7   & 12.1    & 0     & 2.7    & 6.28    & 0      & 2.09  & 6.7   & 15.1   & 23.4  \\
w-l-v3                 & 31.02   & 0      & 2.63   & 10.3    & 0      & 1.9    & 9         & 0        & 2.4      & 6.5      & 0.1     & 8.3      & 16.86     & 0        & 7.06     & 6.5    & 0     & 10.4   & 49.8     & 0       & 1.1    & 11.3    & 9      & 2.6   & 9.1     & 0     & 1.8    & 5.24    & 0      & 1.57  & 6.6   & 17.9   & 25.2  \\
w-l-v3-t               & 30.83   & 0.38   & 2.82   & 13.5    & 0      & 2.6    & 10.4      & 0        & 2.2      & 5.1      & 0.6     & 8.9      & 21.57     & 0        & 9.41     & 6.4    & 0     & 11.3   & 54.9     & 0       & 1.6    & 13.2    & 12.8   & 2     & 10      & 0.1   & 2.2    & 10.99   & 0      & 3.14  & 7.5   & 17.1   & 23.9  \\
w-m                    & 34.59   & 0.19   & 1.88   & 17      & 0.1    & 3.4    & 14.8      & 0        & 2.9      & 6.8      & 0.7     & 10.4     & 16.86     & 0.39     & 10.98    & 6.5    & 0     & 11.2   & 51.5     & 0       & 1.9    & 11.2    & 6.6    & 1.9   & 13.8    & 0     & 2      & 10.47   & 0      & 2.09  & 10.1  & 17.2   & 24    \\
w-m.en                 & 32.71   & 0.19   & 2.63   & 16.4    & 0      & 2.8    & 12.8      & 0        & 3.1      & 6.6      & 0.4     & 10.1     & 17.25     & 0.39     & 9.02     & 6.6    & 0     & 10.3   & 50.1     & 0.8     & 2.4    & 10.2    & 9.4    & 3.5   & 12.5    & 0     & 2.6    & 15.71   & 0      & 2.09  & 6.6   & 16.1   & 25.7  \\
w-m                    & 38.16   & 0      & 2.63   & 28.5    & 0      & 4.8    & 19.6      & 0        & 5.7      & 10.9     & 0.7     & 14.3     & 12.55     & 0.78     & 7.45     & 6.6    & 0     & 12     & 54.3     & 0.3     & 1.9    & 7.7     & 8.3    & 2.2   & 17.5    & 0     & 3.8    & 17.28   & 0      & 1.05  & 10.2  & 18.3   & 27.6  \\
w-s.en                 & 37.22   & 0.38   & 3.38   & 27      & 0      & 5.4    & 18.2      & 0        & 3.9      & 8.2      & 1       & 14.8     & 12.94     & 0.78     & 6.27     & 6.9    & 0     & 11.8   & 57.9     & 0.1     & 1.4    & 6.9     & 11.3   & 2.1   & 15.2    & 0     & 3.4    & 16.23   & 0      & 3.14  & 9.2   & 19.4   & 26.2  \\
w-tiny                 & 36.28   & 2.07   & 4.89   & 26.5    & 0.3    & 10.3   & 33.7      & 0        & 16.8     & 15.3     & 1       & 26.6     & 6.27      & 0.78     & 6.67     & 9.5    & 0     & 16.9   & 32.4     & 1.8     & 2.5    & 0       & 18.2   & 0.5   & 32      & 0     & 12.6   & 33.51   & 0      & 9.95  & 15.6  & 23.7   & 40.3  \\
w-tiny.en              & 34.59   & 2.26   & 4.32   & 29.2    & 0.3    & 12.8   & 30.6      & 0.1      & 14.2     & 12.8     & 1       & 21.4     & 9.02      & 0.78     & 5.1      & 9.3    & 0     & 14.5   & 33       & 1.7     & 3.2    & 0.4     & 25.9   & 1.1   & 25.7    & 0     & 9.6    & 34.03   & 0      & 6.28  & 13.5  & 22     & 36.5 \\

\bottomrule
\end{tabular}
}
\caption{Non-Hallucination error analysis across various datasets and models.The table shows the percentage of Phonetic (P), Oscillation (O), and Language (L) errors for each model evaluated on different datasets. Abbreviations. w – whisper, s – small, m – medium, l – large, t – turbo, dw – distil-whisper, sm4t – seamless, w2v2 – wav2vec2, spllm – SpeechLLM, Qwen2 – Q2A - Qwen2-Audio, SC - Supreme Court.
}
\label{tab:non_hallucination_error_analysis}
\end{table*}

\begin{table}[h!]
\centering
\small
\renewcommand{\arraystretch}{1.3}
\begin{tabularx}{\linewidth}{l X} 
\hline
\textbf{Attribute} & \textbf{Value} \\
\hline
\textbf{Reference} & lufthansa four three nine three descend to flight level two seven zero \\
\textbf{Transcription} & Lufthansa 4393, descent flight level 270. \\
\textbf{WER} & 75.0 \\
\textbf{Hallucination} & No Error \\
\hline
\end{tabularx}
\caption{WER and error category labeled by LLMs for whisper-medium.}
\label{tab:wer_vs_hallucination}
\end{table}



\end{document}